\pdfoutput=1

\documentclass[11pt]{article}
\usepackage[final]{acl}

\usepackage{times}
\usepackage{latexsym}

\usepackage[T1]{fontenc}

\usepackage[utf8]{inputenc}

\usepackage{microtype}

\usepackage{adjustbox}
\usepackage{graphbox}

\usepackage{inconsolata}

\usepackage{graphicx}

\usepackage{bm}
\usepackage{array}
\usepackage{amsfonts}
\usepackage{amsmath}
\usepackage{booktabs}
\usepackage{xspace}
\usepackage{paralist}
\usepackage{algorithm}
\usepackage{colortbl}
\usepackage[noend]{algpseudocode}
\usepackage{multirow}
\usepackage{cuted}
\usepackage{tcolorbox}
\usepackage{caption}

\usepackage{xcolor}
\usepackage[normalem]{ulem}

\tcbuselibrary{breakable}  
\usepackage{multicol}

\definecolor{customGreen}{RGB}{223, 242, 223}
\definecolor{customYellow}{RGB}{250, 249, 214}
\definecolor{customYellowSaturated}{RGB}{255, 225, 119}
\definecolor{customRed}{RGB}{242, 223, 223}
\definecolor{customBlue}{RGB}{214, 224, 250}

\definecolor{firstPageGreen}{RGB}{156, 187, 60}

\definecolor{firstPageBlue}{RGB}{98, 144, 233}
\newcommand{\BlueUL}[1]{{\color{firstPageBlue}\uline{\color{black}#1}}}

\definecolor{firstPageOrange}{RGB}{255, 210, 122}
\newcommand{\OrangeUL}[1]{{\color{firstPageOrange}\uline{\color{black}#1}}}

\newtcolorbox[list inside=prompt,auto counter,number within=section]{prompt}[1][]{
    colbacktitle=black!60,
    coltitle=white,
    fontupper=\footnotesize,
    boxsep=5pt,
    left=0pt,
    right=0pt,
    top=0pt,
    bottom=0pt,
    boxrule=1pt,
    title={#1},
    breakable,
    #1, 
}

\newcolumntype{L}[1]{>{\raggedright\let\newline\\\arraybackslash\hspace{0pt}}m{#1}}
\newcolumntype{C}[1]{>{\centering\let\newline\\\arraybackslash\hspace{0pt}}m{#1}}
\newcolumntype{R}[1]{>{\raggedleft\let\newline\\\arraybackslash\hspace{0pt}}m{#1}}

\newcommand{\pascual}[1]{\textcolor{brown}{(Pascual: #1)}}
\newcommand{\gear}{\textsc{GeAR}\xspace}

\title{\gear: Graph-enhanced Agent for Retrieval-augmented Generation}

\author{{\bf Zhili Shen}$^{\mathbf{\dagger}}$ \quad {\bf Chenxin Diao}$^{\mathbf{\dagger}}$ \quad {\bf Pavlos Vougiouklis}$^{\mathbf{\dagger}}$ \quad {\bf Pascual Merita}$^{\mathbf{\dagger}}$ \\
{\bf Shriram Piramanayagam} \quad {\bf Enting Chen} \quad {\bf Damien Graux} \\ 
\quad {\bf Andre Melo} \quad {\bf Ruofei Lai} \quad {\bf Zeren Jiang} \quad {\bf Zhongyang Li} \\ \quad {\bf Ye Qi}
\quad {\bf Yang Ren} \quad {\bf Dandan Tu} \quad \quad {\bf Jeff Z. Pan}\\ 
Huawei Technologies Co., Ltd.
\\Edinburgh, United Kingdom\\\texttt{\{}{\href{mailto:zhilishen@huawei.com}{\texttt{zhilishen}}}\texttt{,} 
\href{mailto:chenxindiao@huawei.com}{\texttt{chenxindiao}}\texttt{,} \href{mailto:pavlos.vougiouklis@huawei.com}{\texttt{pavlos.vougiouklis}}\texttt{,}
\href{mailto:pascual.merita@h-partners.com}{\texttt{pascual.merita}}\texttt{\}@huawei.com}\\\texttt{\{}{\href{mailto:renyang1@huawei.com}{\texttt{renyang1}}}\texttt{,} 
\href{mailto:jeff.pan@huawei.com}{\texttt{jeff.pan}}\texttt{\}@huawei.com}\\
}

\begin{document}
\maketitle

\begin{abstract}
Retrieval-augmented Generation (RAG) relies on effective retrieval capabilities, yet traditional sparse and dense retrievers inherently struggle with multi-hop retrieval scenarios. In this paper, we introduce \gear, a system that advances RAG performance through two key innovations: \begin{inparaenum}[(i)] \item an efficient graph expansion mechanism that augments any conventional base retriever, such as BM25, and \item an agent framework that incorporates the resulting graph-based retrieval into a multi-step retrieval framework\end{inparaenum}. Our evaluation demonstrates \gear's superior retrieval capabilities across three multi-hop question answering datasets. Notably, our system achieves state-of-the-art results with improvements exceeding $10\%$ on the challenging MuSiQue dataset, while consuming fewer tokens and requiring fewer iterations than existing multi-step retrieval systems. The project page is available at \url{https://gear-rag.github.io}.
\end{abstract}

\renewcommand{\thefootnote}
{\fnsymbol{footnote}}
\setcounter{footnote}{2}
\footnotetext{The authors contributed equally to this work.}
\renewcommand{\thefootnote}
{\arabic{footnote}}
\setcounter{footnote}{0}

\section{Introduction}
\label{sec:intro}
Retrieval-augmented Generation (RAG) has enhanced the performance of Large Language Models (LLMs) \cite{OpenAI2024} in Question Answering (QA) tasks \cite{Lewis2020}. While effective for simple queries, multi-hop QA presents a more complex challenge, requiring reasoning across several passages or documents. Consider the example in Table~\ref{tab:qa_example}, where finding the correct answer requires building a $3$-hop reasoning chain starting from the question's main entity (i.e. ``Stephen Curry''). A base retriever cannot, by design, retrieve all necessary information in a single step.

\begin{table}[ht]
    \small
    \centering
    \begin{tabular}{p{7.1cm}}
    \toprule
    In what year did Stephen Curry's father join the team from which he started his college basketball career? \\\midrule
    \includegraphics[width=7.2cm]{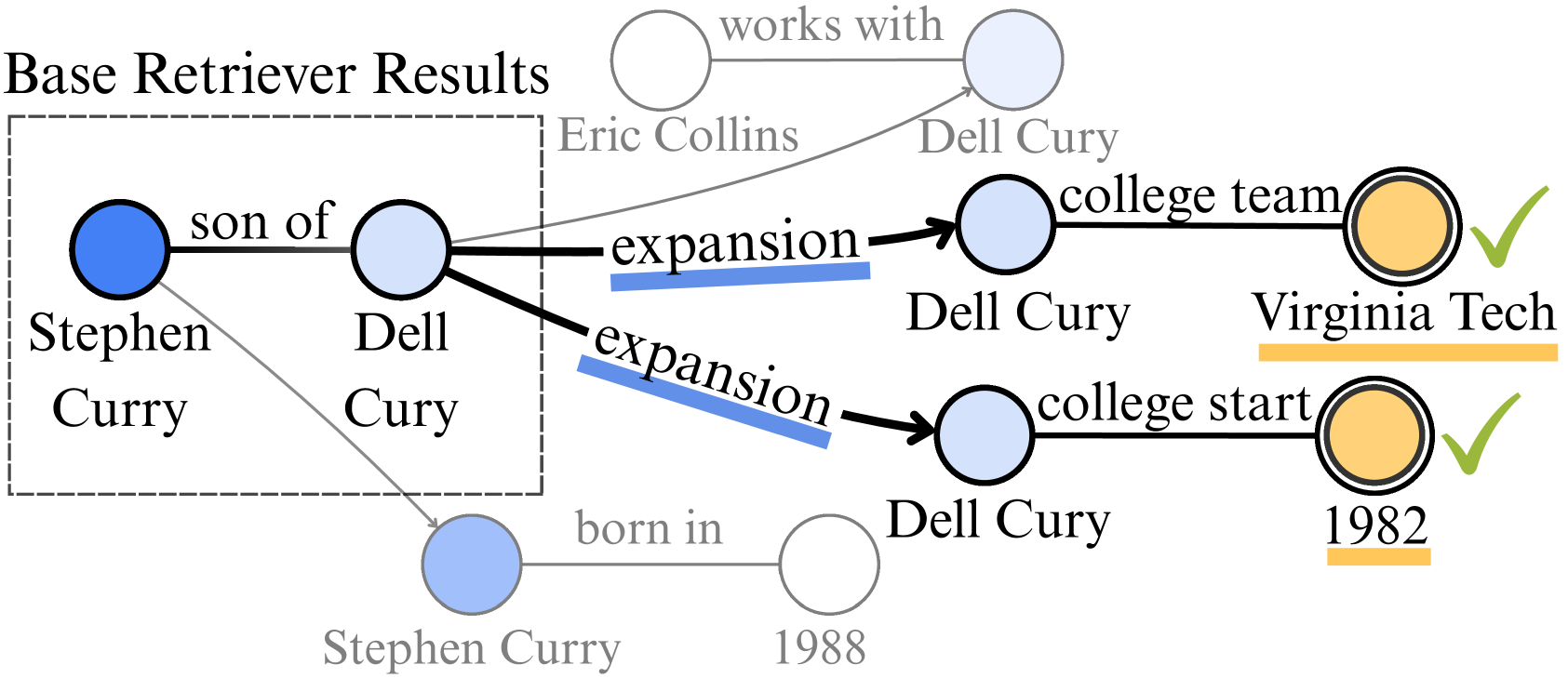} \\\bottomrule
    \end{tabular}
    \caption{A \textbf{motivating example} of a multi-hop question where a base retriever cannot, by design, retrieve all necessary information in a single step. Graph \BlueUL{expansion} (see~\S\ref{subsec:diverse_triple_beam_search}), which we incorporate within \gear, enables retrieval of subsequent hops and guides the system toward the correct \OrangeUL{answer} without using an LLM.}
    \label{tab:qa_example}
\end{table}

To address these complex reasoning requirements, researchers have increasingly turned to graph-based approaches \cite{Fang2024,Li2024,Edge2024,Gutierrez2024,Liang2024}. By extracting entities, atomic facts, or semantic triples \cite{Li2024, Fang2024, Gutierrez2024}, these graphs can establish more direct pathways for multi-hop reasoning. For instance, HippoRAG extracts triples from passages to form a knowledge graphs and employs a variant of PageRank for passage retrieval \cite{Gutierrez2024}. GraphReader uses an LLM agent with graph-navigating operations
to explore the resulting graph structure \cite{Li2024}. TRACE relies on an LLM to iteratively select triples for constructing reasoning chains, which then either ground answer generation, or filter irrelevant documents from initially retrieved results \cite{Fang2024}. However, recursively prompting LLMs to traverse graphs remains computationally expensive, particularly as search spaces expand.


In this paper, we present \gear, a \underline{\textbf{G}}raph-\underline{\textbf{e}}nhanced \underline{\textbf{A}}gent for \underline{\textbf{R}}etrieval-augmented generation. During the offline stage, we \textit{align} passages with their extracted triples to create interconnected indices. This alignment allows passages to be connected through graphs of triples. \gear features a graph-based passage retrieval component called SyncGE. Unlike previous approaches that require expensive LLM calls for graph exploration, our method uses an LLM only to locate initial nodes (triples) and then employs a generic semantic model to expand the triple sub-graph by exploring diverse triple beams. Additionally, \gear uses multi-hop context retrieved by SyncGE to construct a memory that summarises information for multi-step retrieval.

Our work refines the neurobiology-inspired paradigm proposed by \citeauthor{Gutierrez2024} by modeling the communication between the hippocampus and neocortex during episodic memory formation. In our design, an array of \textit{proximal triples} functions as a memory gist learned through the hippocampus within one or a few shots (iterations), which is then projected back to the neocortex for later recall stages \cite{Hanslmayr2016,Griffiths2019}. We highlight the complementary potential between our graph retrieval approach and an LLM, which, within our system, emulates the synergy between the hippocampus and neocortex, offering insights from a biomimetic perspective.

We evaluate the retrieval performance of \gear on three multi-hop QA benchmarks: MuSiQue, HotpotQA, and 2WikiMultihopQA. \gear pushes the state of the art, achieving significant improvements in both single- and multi-step retrieval settings, with gains exceeding $10\%$ on the most challenging MuSiQue dataset. Furthermore, we demonstrate that our framework can address multi-hop questions in fewer iterations while consuming significantly fewer LLM tokens. Even with a single iteration, \gear offers a more efficient alternative to other iterative retrieval methods, such as HippoRAG w$/$ IRCoT. Our contributions can be summarised as follows:

\begin{itemize}
\item We introduce a novel graph-based retriever, SyncGE, which leverages an LLM for locating initial nodes for graph exploration and subsequently expands them by diversifying beams of triples that link multi-hop passages.
\item We incorporate this graph retrieval method within an LLM-based agent framework, materialising \gear, achieving state-of-the-art retrieval performance across three datasets.
\item We conduct comprehensive experiments showcasing the synergetic effects between our proposed graph-based retriever and the LLM within the \gear framework.
\end{itemize}

\section{Related Work}

Our work draws inspiration from two branches of research: \begin{inparaenum}[(i)]\item retrieval-augmented models for QA and \item multi-hop QA using combinations of LLMs with graphical structures\end{inparaenum}.

\subsection{Retrieval-augmented Models for QA}

\citeauthor{Lewis2020} first showcased the benefits of augmenting language models' input context with relevant passages. 
Recent work by \citeauthor{Wang2023a} and \citeauthor{Shen2024} explores query expansion approaches, generating pseudo-documents from language models to expand the content of original queries. Subsequent frameworks, beginning with IRCoT, have investigated the interleaving of retrieval and prompting steps, allowing each step to iteratively guide and refine the other \cite{Trivedi2023,Jiang2023,Su2024}.

\subsection{Multi-hop QA with LLMs and Graphs}

Several architectures have introduced an offline indexing phase to form hierarchical passage summaries \cite{Chen2023,Sarthi2024,Edge2024}. However, summarisation must be repeated when adding new data, making knowledge base updates computationally expensive. Recent approaches have leveraged structured knowledge to address multi-hop QA challenges with LLMs~\cite{Park2023,Shen2024a,Li2024,Gutierrez2024,Liang2024,Wang2024}. GraphReader, TRACE and HippoRAG propose offline methods for extracting entities and atomic facts or semantic triples from passages \cite{Li2024,Fang2024,Gutierrez2024}.
This allows chunks containing the same or neighbouring entities to construct a graph of indexed passages. TRACE relies on an LLM to iteratively select triples for reasoning chains, which ground answer generation or filter retrieved results. However, its search space is limited by pre-filtered candidate lists for each query. \citeauthor{Li2024} employ an LLM agent that selects from a set of predefined actions to traverse knowledge graph nodes in real time given an input question. \citeauthor{Liang2024} later introduced additional graph standardisation, including instance-to-concept linking and semantic relation completion. However, this approach heavily depends on associating triples with pre-defined concepts for logical form-based retrieval.

HippoRAG leverages an alignment of passages and extracted triples to retrieve passages based on the Personalized PageRank algorithm \cite{Gutierrez2024}.
While achieving improvements for single- and multi-step retrieval (when coupled with IRCoT \cite{Trivedi2023}), it remains agnostic to the semantic relationships of extracted triples. In this paper, we leverage a similar alignment of passages and extracted triples but introduce a new graph-based retrieval framework that uses a small semantic model for exploring multi-hop relationships. Our framework considers the contributions of all triple elements participating in reasoning chains, offering a more robust solution for associating questions with triple reasoning chains.

\section{Preliminaries}
\label{sec:preliminaries}

Let $\mathbf{C} = \left \{c_1, c_2, \ldots, c_C \right \}$ be an index of passages and $\mathbf{T} = \left \{t_1, t_2, \ldots,t_T: t_j = \left ( s_j, p_j, o_j \right ) \right \}$ be another index representing a set of triples associated with the passages in $\mathbf{C}$ s.t. $\forall t_j \in \mathbf{T} \exists! \, c_i \in \mathbf{C}$, where $s_j$, $p_j$ and $o_j$ the respective subject, predicate and object of the $j$-th triple. In this setup, each triple is uniquely linked to exactly one passage, and a passage can potentially be associated with multiple triples.

Given an input query $\mathbf{q}$ and an index of interest $\mathbf{R} = \left \{r_1, \ldots, r_R \right \}$, retrieving items from $\mathbf{R}$ relevant to $\mathbf{q}$ can be achieved by using a base retrieval function $h^k_{\text{base}}\left( \mathbf{q}, {\mathbf{R}}\right ) \subseteq \mathbf{R}$ that returns a ranked list of $k$ items from $\mathbf{R}$ in descending order, according to a retrieval score. BM25 or a conventional dense retriever can serve as a base retrieval function, without requiring any multi-hop capabilities.

\begin{figure}[thbp]
  \includegraphics[width=\columnwidth]{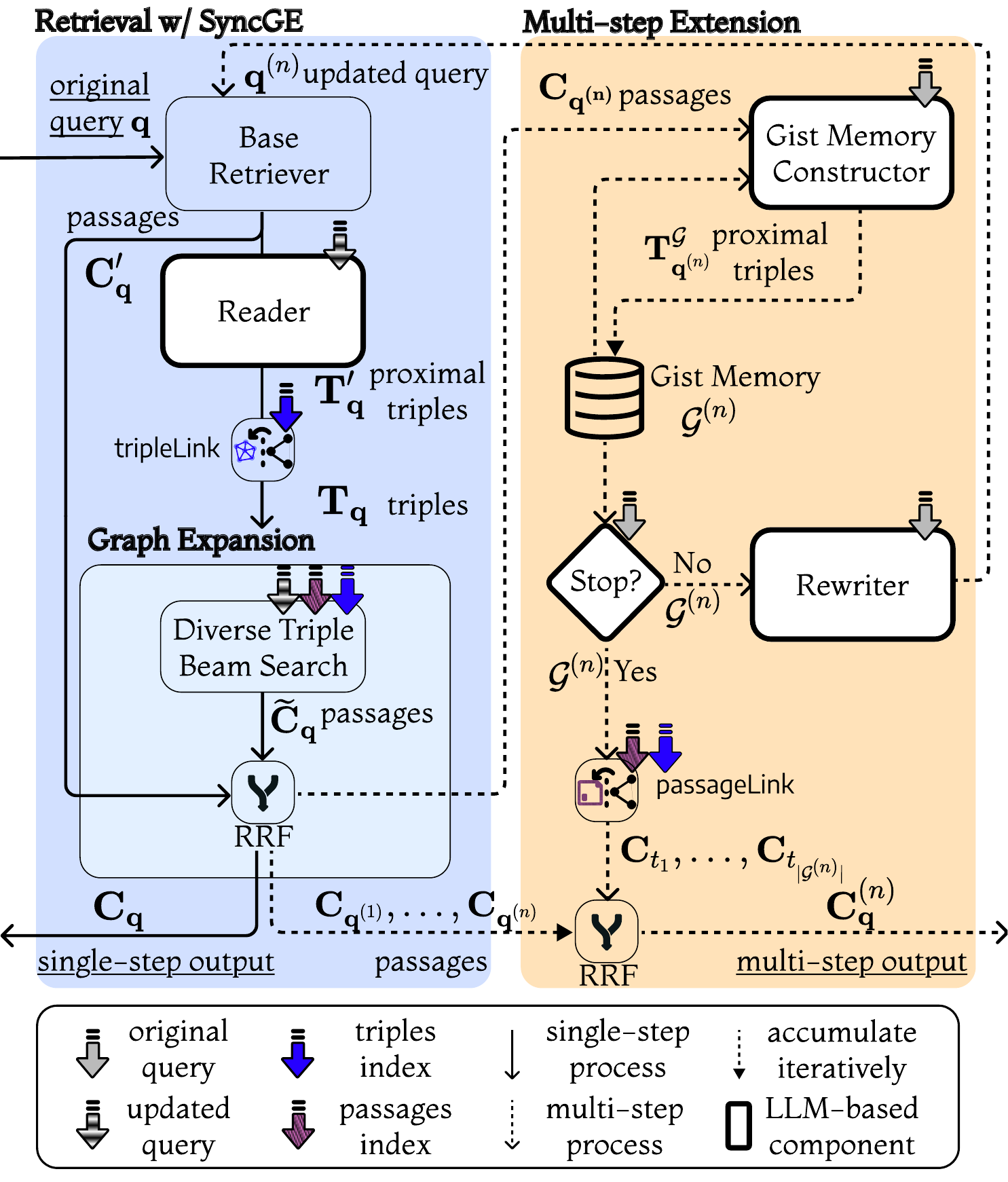}
  \caption{\label{fig:system_diagram}\textbf{System Architecture}. The left section (blue) depicts SyncGE (detailed in~\S\ref{sec:graph_retrieval}), while the the right section (orange), completes the \gear system (described in~\S\ref{sec:gear}). A visual step-by-step walk-through of the system is presented in Appendix \ref{appendix_sec:walkthrough_example}.}
\end{figure}

Our goal is to retrieve relevant passages from $\mathbf{C}$ that enable a retrieval-augmented model to answer multi-hop queries \cite{Lewis2020}. To this end, we introduce \gear, which is a graph-enhanced framework of retrieval agent (see Figure~\ref{fig:system_diagram}). 

\section{Retrieval with Synchronised Graph Expansion}
\label{sec:graph_retrieval}

\def\Tqinit{\mathbf{T}_\mathbf{q}}

Given an input query $\mathbf{q}$, let $\mathbf{C}_\mathbf{q}' = h^k_{\text{base}}\left( \mathbf{q}, {\mathbf{C}}\right )$  be a list of passages returned by the base retriever.
Given this list, $\mathbf{C}_\mathbf{q}'$, our goal is to derive relevant multi-hop contexts (passages) by retrieving a sub-graph of triples that interconnect their source passages. Two challenges arise in materialising such sub-graph retrieval: \begin{inparaenum}[(i)]\item locating initial triples (i.e. starting nodes) $\Tqinit$, and \item expanding the graph from these initial triples while reducing the search space\end{inparaenum}. The following sections address these challenges within \gear.

\subsection{Knowledge Synchronisation}
\label{subsection:knowledge_syncro}
\def\linkTriple{\texttt{tripleLink}}

We describe a knowledge \textbf{Sync}hronisation (\textbf{Sync}) process to locate initial nodes for graph expansion. We first employ an LLM to \texttt{read} $\mathbf{C}_\mathbf{q}'$ (see Appendix~\ref{subsec:online_retrieval_prompts}) and summarise knowledge triples that can support answering the current query $\mathbf{q}$:
\begin{align}
    \mathbf{T}_\mathbf{q}' = \texttt{read}\left (\mathbf{C}_\mathbf{q}', \mathbf{q}\right ).
    \label{eq:proximal_read}
\end{align}

$\mathbf{T}_\mathbf{q}'$ is a collection of triples to which we refer as \textit{proximal triples}. Initial nodes $\Tqinit$ for graph expansion are identified by linking each triple in $\mathbf{T}_\mathbf{q}'$ to a triple in $\mathbf{T}$, using the \linkTriple{} function:
\begin{align}
    \Tqinit =\left \{t_i | t_i = \linkTriple(t_i') ~ \forall t_i' \in \mathbf{T}_\mathbf{q}'\right \}.
\end{align}
The implementation of \linkTriple{} can vary. However, in this paper we consider it to be simply retrieving the most similar triple from $\mathbf{T}$ using $h^1_{\text{base}}\left( t_i', {\mathbf{T}}\right) \forall t_i' \in \mathbf{T}_\mathbf{q}'$.

\begin{algorithm}[ht]
\textbf{Input:} $\mathbf{q}$: query \\
\hspace*{3em} $b$: beam size \\
\hspace*{3em} $l$: maximum length \\
\hspace*{3em} $\mathrm{score}(\cdot, \cdot)$: scoring function \\
\hspace*{3em} $\{t_1, t_2, \ldots, t_n\}$: initial triples \\
\hspace*{3em} $\gamma$: hyperparameter for diversity

\begin{algorithmic}[1]
\State $B_0 \gets [\;]$
\For{$t \in \{t_1, t_2, ..., t_n\}$}
    \State $s \gets \mathrm{score}(\mathbf{q}, [t])$
    \State $B_0.\mathrm{add}(\langle s, [t] \rangle)$
\EndFor

\State $B_0 \gets \mathrm{top}(B_0, b)$

\For{$i \in \{1, \dots, l - 1\}$}
    \State $B \gets [\;]$
    
    \For{$\langle s, T \rangle \in B_{i-1}$}
        \State $V \gets [\;]$

        \For{$t \in \mathrm{get\_neighbours}(T.\mathrm{last}())$}
            \If{$\mathrm{exists}(t, B_{i-1})$}
                \State \textbf{continue}
            \EndIf
            
            \State $s' \gets s + \mathrm{score}(\mathbf{q}, T \circ t)$ ~ \texttt{\# concat} 
            \State $V.\mathrm{add}(\langle s', T \circ t \rangle)$
        \EndFor

        \State $\mathrm{sort}(V, \mathrm{descending})$

        \For{$n \in \{0, \dots, V.\mathrm{length()} - 1\}$}
            \State $\langle s', T \circ t \rangle \gets V[n]$
            \State $s' \gets s' \times e^{- \frac{\mathrm{min}(n, \gamma)}{\gamma}}$
            \State $B.\mathrm{add}(\langle s', T \circ t \rangle)$
        \EndFor
        
    \EndFor
    \State $B_i \gets \mathrm{top}(B, b)$
    
\EndFor

\State \Return $B_i$
\end{algorithmic}

\caption{Diverse Triple Beam Search}
\label{alg:beam_search}
\end{algorithm}

\subsection{Diverse Triple Beam Search}
\label{subsec:diverse_triple_beam_search}
We borrow the idea of constructing reasoning triple chains \cite{Fang2024} for expanding the graph, and present a retrieval algorithm: \textit{Diverse Triple Beam Search} (see Alg.~\ref{alg:beam_search}). 

We maintain top-$b$ sequences (beams) of triples and the scores at each step are determined by a scoring function. In this paper, we focus on leveraging a dense embedding model to compute the cosine similarity between embeddings of the query and a candidate sequence of triples, leaving other implementations of the scoring function for future work (see Section~\ref{sec:limitations}).

Considering all possible triple extensions at each step, in a Viterbi decoding fashion, would be intractable due to the size of $\mathbf{T}$. Consequently, we define the neighbourhood of a triple as the set of triples with shared head or tail entities (i.e. $\mathrm{get\_neighbours}$ in Alg.~\ref{alg:beam_search}). During each expansion step, we only consider neighbours of the last triple in the sequence, and avoid selecting previously visited triples (i.e. $\mathrm{exists}$ in Alg.~\ref{alg:beam_search}) to further reduce the search space.

While regular beam search can reduce the search space, it is prone to producing high-likelihood sequences that differ only slightly from one another \cite{Ippolito2019, Vijayakumar2018}. Our algorithm increases the diversity across beams to improve the recall for retrieval. In detail, for each beam, we sort candidate sequences extended from that beam in descending order, and weight their scores based on their relative positions. Candidate sequences that are ranked lower, within a beam, will receive smaller weights. Consequently, the resulting top-$b$ beams at each step are less likely to share the same starting sequence. 

The top-$b$ returned sequences are flattened in a breadth-first order. Each triple in the resulting list is then mapped to its source passage. This alignment between triples and passages is described in more detail in Section~\ref{sec:preliminaries}. Let $\widetilde{\mathbf{C}}_\mathbf{q}$ be the list of unique passages after alignment. The output of our graph expansion is then given by the Reciprocal Rank Fusion (RRF) \cite{Cormack2009} of $\widetilde{\mathbf{C}}_\mathbf{q}$ and the initial $\mathbf{C}_\mathbf{q}'$ list of passages :
\begin{align}
    \mathbf{C}_{\mathbf{q}} = \mathrm{RRF}\left(\widetilde{\mathbf{C}}_\mathbf{q}, \mathbf{C}_\mathbf{q}'\right ).
\end{align}
We refer to this
method for retrieving passages as \underline{\smash{\textbf{Sync}}}ronised \underline{\smash{\textbf{G}}}raph \underline{\smash{\textbf{E}}}xpansion (\textbf{SyncGE}).

\section{Multi-step Extension}
\label{sec:gear}

We further present an agentic framework that models a human-like information-seeking process through multi-turn interactions with the graph-enhanced retriever. The resulting agent is referred to as \gear. We focus on:
\begin{itemize}
\item maintaining a gist memory of proximal knowledge obtained throughout the different steps 
\item incorporating a similar synchronisation process 
that summarises retrieved passages in proximal triples to be stored in this multi-turn gist memory
\item determining if additional steps are needed for answering the original input question
\end{itemize}
Within this multi-turn setting, the original input question $\mathbf{q}$ is iteratively decomposed into simpler queries: $\mathbf{q}^{(1)}, \ldots, \mathbf{q}^{(n)}$, where $\mathbf{q}^{(1)} = \mathbf{q}$ and $n \in \mathbb{N}$ represents the number of the current step.
For each query $\mathbf{q}^{(n)}$, we use the graph retrieval method introduced in Section~\ref{sec:graph_retrieval} in order to retrieve relevant passages $\mathbf{C}_{\mathbf{q}^{(n)}}$.

\subsection{Gist Memory Constructor}
\label{subsec:gist_memory_constructor}
To facilitate the multi-step capabilities of our agent, we introduce a \textit{gist memory}, $\mathcal{G}^{(n)}$, which is used for storing knowledge as an array of proximal triples. At the beginning of the first iteration, the gist memory is empty. During the $n$-th iteration, similar to the knowledge synchronisation module explained in Section~\ref{subsection:knowledge_syncro}, we employ an LLM to read a collection of retrieved paragraphs $\mathbf{C}_{\mathbf{q}^{(n)}}$ and summarise their content with proximal triples:

\begin{align}
\mathbf{T}_{\mathbf{q}^{(n)}}^{\mathcal{G}} = 
\begin{cases} 
    \texttt{read}\left(\mathbf{C}_{\mathbf{q}^{(n)}}, \mathbf{q} \right), & \text{if } n = 1 \\
    \texttt{read}\left(\mathbf{C}_{\mathbf{q}^{(n)}}, \mathbf{q}\textcolor{blue}{, \mathcal{G}^{(n-1)}} \right), & \text{if } n \geq 2
\end{cases}
\label{eq:proximal_read_agent}
\end{align}

Apart from the first iteration where Eq.~\ref{eq:proximal_read} and ~\ref{eq:proximal_read_agent} are identical, the inclusion of the memory in the \texttt{read} operation differentiates the construction of proximal triples produced at the subsequent steps compared to the ones from Eq.~\ref{eq:proximal_read}. $\mathcal{G}^{(n)}$ maintains the aggregated content of proximal triples s.t. 
\begin{align}
\mathcal{G}^{(n)} = \left[ \mathbf{T}_{\mathbf{q}^{(1)}}^{\mathcal{G}}  \circ \cdots \circ \mathbf{T}_{\mathbf{q}^{(n)}}^{\mathcal{G}} \right],
\end{align}where $\circ$ defines the concatenation operation. The triple memory serves as a concise representation of all the accumulated evidence, up to the $n$-th step. 

We believe the process introduced by the \texttt{read} step along with the information storage paradigm served by the gist memory, aligns well with the communication between the hippocampus and neocortex. The combination of the two establishes the synergetic behaviour between our graph retriever and the LLM that we seek to achieve within \gear.

\subsection{Reasoning for Termination}
\label{subsec:reasoning}
After updating $\mathcal{G}^{(n)}$, we assess whether it contains sufficient evidence to answer the original question through an LLM reasoning step:
\begin{align}
\mathbf{a}^{(n)}, \mathbf{r}^{(n)}   = \texttt{reason}(\mathcal{G}^{(n)}, \mathbf{q}),
\end{align}
where $\mathbf{a}^{(n)}$ denotes the query's answerability given the 
evidence in $\mathcal{G}^{(n)}$, and $\mathbf{r}^{(n)}$ represents the relevant reasoning. When the query is deemed answerable, the system concludes its iterative process.

\subsection{Query Re-writing}
\label{subsec:rewriting}
The query re-writing process leverages an LLM that incorporates three key inputs: the original query $\mathbf{q}$, the accumulated memory, and crucially, the reasoning output $\mathbf{r}^{(n)}$ from the previous step. This process can be formally expressed as:
\begin{align}
\mathbf{q}^{(n+1)} = \texttt{rewrite}\left (\mathcal{G}^{(n)}, \mathbf{q}, \mathbf{r}^{(n)} \right),
\end{align}
where $\mathbf{q}^{(n+1)}$ represents the updated query, which serves as input for the retriever in the next iteration.\\
\subsection{After Termination}
\gear aims to return a single ranked list of passages. Given the final gist memory $\mathcal{G}^{(n)}$ upon termination, we link each proximal triple in $\mathcal{G}^{(n)}$ to a list of passages as follows:
\begin{align}
    \mathbf{C}_{t_j} = \texttt{passageLink}\left(t_j, k\right),
\end{align}
where $j \in \left \{1, \dots, \vert\mathcal{G}^{(n)}\vert \right \}$. Similar to \texttt{tripleLink}, \texttt{passageLink} is implemented by retrieving passages with a triple as the query (see Appendix~\ref{appendixpara:passage_link}). The final list of passages returned by \gear is the RRF of the resulting linked passages and passages retrieved across iterations:
\begin{align}
\mathbf{C}_\mathbf{q}^{(n)} = \mathrm{RRF}\big(&\mathbf{C}_{t_1}, \ldots,\mathbf{C}_{t_{\vert\mathcal{G}^{(n)}\vert}}, \nonumber\\
    &\mathbf{C}_{\mathbf{q}^{(1)}}, \ldots, \mathbf{C}_{\mathbf{q}^{(n)}} \big).
\end{align}

All relevant prompts for the \texttt{read}, \texttt{reason} and \texttt{rewrite} steps are provided in Appendix~\ref{subsec:online_retrieval_prompts}.

\section{Experimental Setup}
We evaluate our framework on three open-domain multi-hop QA datasets: \textbf{MuSiQue} \cite{Trivedi2022}, \textbf{HotpotQA} \cite{Yang2018}, and \textbf{2WikiMultiHopQA} (2Wiki) \cite{Ho2020}. For MuSiQue and 2Wiki, we use the data provided in the IRCoT paper \cite{Trivedi2023}
which includes the full corpus, while for HotpotQA, we follow the same setting as HippoRAG \cite{Gutierrez2024} to limit experimental costs. More details are provided in Appendix \ref{appendix:dataset_stats}.

We measure both retrieval and QA performance, with our primary contributions focused on the retrieval component. For retrieval evaluation, we use Recall@$k$ (R@$k$) for $k \in \left \{5, 10, 15\right \}$, showing the percentage of questions where the correct entries are found within the top-$k$ retrieved passages. We include an analysis about the selected recall ranks in Appendix \ref{appendix:reasoning_behind_retrieval_metrics}. Following standard practices, QA performance is evaluated with Exact Match (EM) and F1 scores \cite{Trivedi2023}.

\begin{table*}[t]
\small
\centering
\small
\begin{tabular}{@{}l@{\hspace{2pt}}lccccccccc@{}}
\toprule
& \multirow{2.5}{*}{\textbf{Retriever}} & \multicolumn{3}{c}{\textbf{MuSiQue}} & \multicolumn{3}{c}{\textbf{2Wiki}} & \multicolumn{3}{c}{\textbf{HotpotQA}}\\ 
\cmidrule{3-11}
& & R@5 & R@10 & R@15 & R@5 & R@10 & R@15 & R@5 & R@10 & R@15 \\ \midrule
\multirow{11}{*}{\parbox{2cm}{\textbf{Single-step\\Retrieval}}}
& ColBERTv2 & $39.4$ & $44.8$ & $47.7$ & $59.1$ & $64.3$ & $66.2$ & $79.3$ & $87.1$ & $90.1$ \\
& HippoRAG & $41.0$ & $47.0$ & $51.4$ & $\mathbf{75.1}$ & $\mathbf{83.2}$ & $\mathbf{86.4}$ & $79.8$ & $89.0$ & $92.4$ \\ 
& BM25 & $33.8$ & $38.5$ & $41.3$ & $59.5$ & $62.7$ & $64.1$ & $74.2$ & $83.6$ & $86.3$ \\ 
& \hspace{2mm} + NaiveGE & $37.5$ & $45.5$ & $48.4$ & $65.0$ & $70.7$ & $71.8$ & $79.1$ & $89.1$ & $91.9$ \\ 
& \hspace{2mm} + SyncGE & $\underline{44.7}$ & $\underline{52.6}$ & $\underline{57.4}$ & $70.5$ & $76.1$ & $79.3$ & $\underline{87.4}$ & $\underline{93.0}$ & $\underline{94.0}$ \\ 
& SBERT & $31.1$ & $37.9$ & $41.6$ & $41.2$ & $48.1$ & $51.5$ & $72.1$ & $79.3$ & $84.0$ \\
& {\hspace{2mm} + NaiveGE} & $32.2$ & $41.4$ & $45.4$ & $45.1$ & $54.0$ & $57.3$ & $76.1$ & $84.7$ & $88.8$ \\
& \hspace{2mm} + SyncGE & $41.6$ & $51.3$ & $54.2$ & $54.8$ & $64.9$ & $70.7$ & $84.1$ & $89.6$ & $92.8$ \\ 
& Hybrid & $39.9$ & $46.3$ & $49.1$ & $60.0$ & $65.8$ & $66.6$ & $77.8$ & $85.8$ & $89.7$ \\
& \hspace{2mm} + NaiveGE & $41.8$ & $49.4$ & $53.0$ & $63.0$ & $70.8$ & $72.6$ & $80.6$ & $89.4$ & $92.7$ \\
& {\hspace{2mm} + SyncGE} & $\mathbf{48.7}$ & $\mathbf{57.7}$ & $\mathbf{61.2}$ & $\underline{72.6}$ & $\underline{80.9}$ & $\underline{82.4}$ & $\mathbf{87.4}$ & $\mathbf{93.3}$ & $\mathbf{95.2}$ \\ 
\midrule
\multirow{4}{*}{\parbox{2cm}{\textbf{Multi-step}\\ \textbf{Retrieval}}}
& IRCoT (BM25) & $46.1$ & $\underline{54.9}$ & $57.9$ & $67.9$ & $75.5$ & $76.1$ & $87.0$ & $92.6$ & $92.9$ \\
& IRCoT (ColBERTv2) & $47.9$ & $54.3$ & $56.4$ & $60.3$ & $86.6$ & $69.7$ & $86.9$ & $92.5$ & $92.8$ \\
& HippoRAG w$/$ IRCoT
& $\underline{48.8}$ & $54.5$ & $\underline{58.9}$ & $\underline{82.9}$ & $\underline{90.6}$ & $\underline{93.0}$ & $\underline{90.1}$ & $\underline{94.7}$ & $\underline{95.9}$ \\
& \gear & $\mathbf{58.4}$ & $\mathbf{67.6}$ & $\mathbf{71.5}$ & $\mathbf{89.1}$ & $\mathbf{95.3}$ & $\mathbf{95.9}$ & $\mathbf{93.4}$ & $\mathbf{96.8}$ & $\mathbf{97.3}$ \\ \bottomrule
\end{tabular}
 \caption{\textbf{Retrieval performance} for single- and multi-step retrievers on MuSiQue, 2Wiki, and HotpotQA. Results are reported using Recall@$k$ (R@$k$) metrics for $k \in \left \{5, 10, 15\right \}$.}
 
 \label{tab:recall_main_table}
\end{table*}

\subsection{Baselines}

We evaluate \gear against strong, multi-step baselines, including IRCoT \cite{Trivedi2023} and HippoRAG w$/$ IRCoT \cite{Gutierrez2024}, which, similar to our framework, combines graph retrieval and a multi-step agent. To demonstrate our graph retriever's (i.e. SyncGE) benefits, we evaluate it against several stand-alone, single-step retrievers: \begin{inparaenum}[(i)]\item BM25, \item Sentence-BERT (SBERT), \item a hybrid approach combining BM25 and SBERT through RRF and \item HippoRAG\end{inparaenum}. Following \citeauthor{Gutierrez2024}, we refer to the single-step setup when multiple LLM iterations are not supported.

\subsection{Implementation Details}
\label{subsec:implementation_details}
We reproduce HippoRAG and IRCoT using the code provided by \citeauthor{Gutierrez2024}. To ensure fair comparisons, we employ GPT-4o mini (\texttt{gpt-4o-mini-2024-07-18}) for all methods that require an LLM as well as their corresponding triple extraction. The temperature is set to 0. Our triple extraction prompt (in Appendix \ref{sec:offline_prompts}) is inspired by \citeauthor{Gutierrez2024}. We run QA experiments using the prompts provided in Appendix~\ref{subsec:online_qa_prompts}.


In addition to our proposed SyncGE, we consider a more \textit{naive} implementation of GE (i.e. NaiveGE) to evaluate the performance when no LLM is involved and further demonstrate the effectiveness of synchronisation. In NaiveGE, we input all triples associated with $\mathbf{C}_\mathbf{q}'$ (see Section~\ref{sec:graph_retrieval}) for diverse triple beam search. Comprehensive implementation details are provided in Appendices ~\ref{appendix_sec:detailed_implementation_details}--\ref{appendix_sec:hipporag_results_original_prompt}. Additionally, Appendix \ref{appendix_sec:ablation_studies} provides more experiments evaluating \gear with varying configurations.

\section{Results}

\paragraph{\gear demonstrates state-of-the-art performance in multi-step retrieval}
The multi-step results in Table \ref{tab:recall_main_table} show that our agent-based approach to multi-step retrieval is highly effective, achieving state-of-the-art results across all datasets. While we see significant improvements on saturated datasets like 2Wiki and HotpotQA, \gear especially excels on MuSiQue, delivering performance gains of over 10\% compared to competitors.

\begin{figure}[t]
\includegraphics[width=\columnwidth]{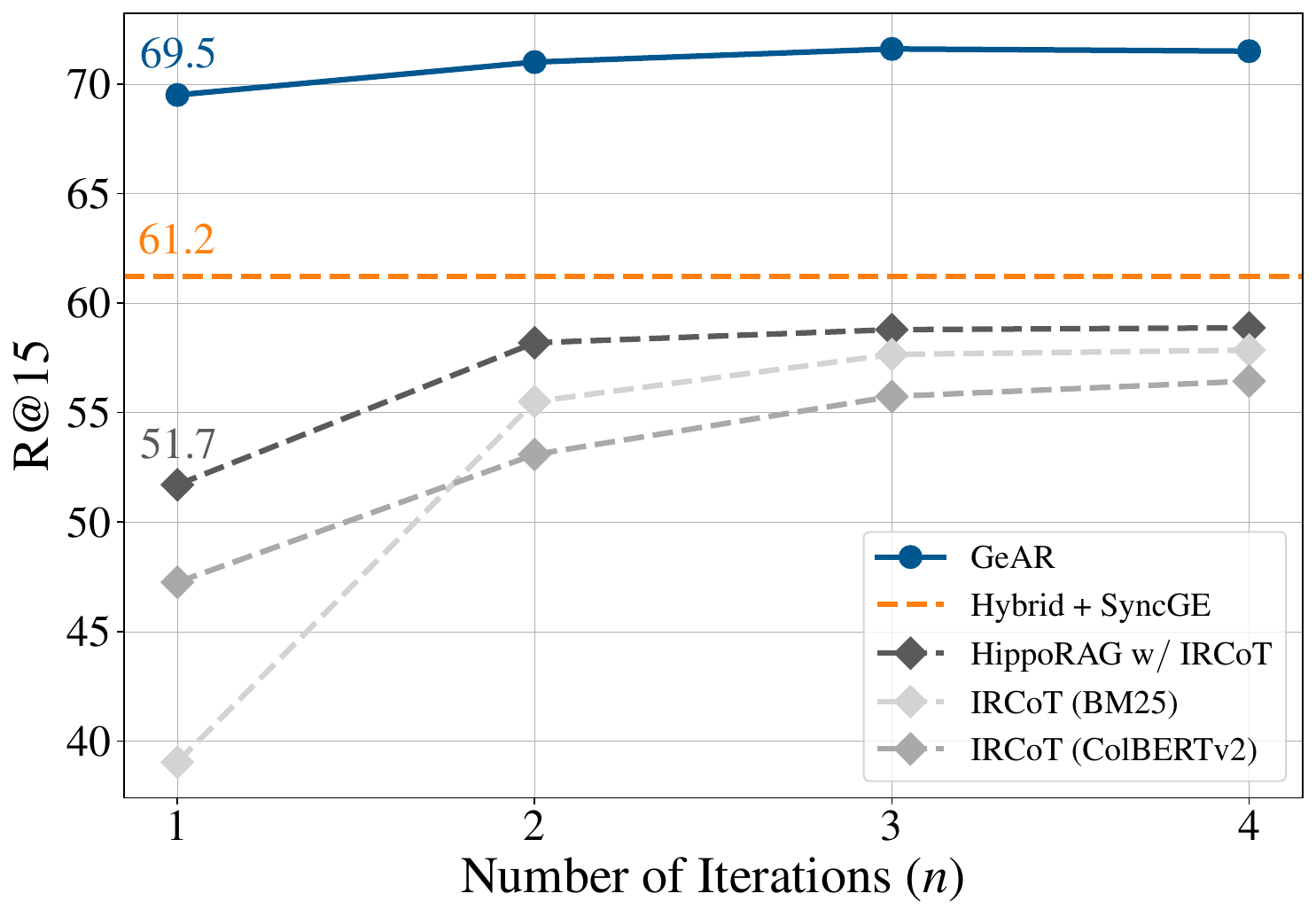}
\caption{\label{fig:recall_across_iterations}R@15 over $4$ iterations on MuSiQue. Recall is computed at each iteration using the cumulative set of retrieved documents, with prior recall values carried forward for questions that terminated in earlier iterations. The horizontal line indicates the single-step performance of Hybrid + SyncGE.}
\end{figure}

\paragraph{SyncGE contributes to state-of-the-art performance in single-step retrieval}
As shown in the single-step section of Table \ref{tab:recall_main_table}, our proposed Hybrid + SyncGE method achieves state-of-the-art single-step retrieval performance on both MuSiQue and HotpotQA datasets. We observe consistent improvements using NaiveGE and SyncGE, outperforming HippoRAG in many setups regardless of the base retriever (i.e. sparse, dense or hybrid). Most notably, Hybrid + SyncGE surpasses HippoRAG by up to $9.8\%$ at R@15 on MuSiQue.

\paragraph{Higher recall leads to higher QA performance}Our analysis shows a positive correlation between recall and QA performance, aligning with the results of prior works \cite{Gutierrez2024}. As shown in Table~\ref{tab:em_and_f1}, \gear achieves the highest EM and F1 scores. A closer examination reveals interesting insights. Taking MuSiQue as an example, \gear shows a $21\%$ relative improvement in R@15 compared to HippoRAG w$/$ IRCoT, while achieving a $37\%$ relative improvement in both EM and F1 scores. Similarly to Table~\ref{tab:recall_main_table}, SyncGE outperforms HippoRAG on MuSiQue and HotpotQA.

\begin{table}[thbp]
\small
\setlength{\tabcolsep}{3pt}
  \centering
  \small
\begin{tabular}{L{2.3cm}cccccc}
\toprule \multirow{2.5}{*}{\textbf{Retriever}} & \multicolumn{2}{c}{\textbf{MuSiQue}} & \multicolumn{2}{c}{\textbf{2Wiki}} & \multicolumn{2}{c}{\textbf{HotpotQA}} \\ \cmidrule{2-7} & EM  & F1 & EM & F1 & EM & F1 \\ \midrule
\rowcolor[gray]{0.95} No Passages & $2.6$ & $12.5$& $17.2$ & $27.9$ & $19.5$ & $34.3$ \\
\rowcolor[gray]{0.95} Gold Passages & $36.6$ & $59.2$ & $54.4$ & $70.3$ & $55.0$ & $75.9$  \\ \midrule
Hybrid + SyncGE & $14.0$ & $\underline{27.1}$ & $38.0$ & $50.2$ & $45.0$ & $63.4$ \\
HippoRAG & $8.2$ & $18.2$ & $39.8$ & $51.8$ & $40.1$ & $57.6$ \\ \midrule
IRCoT (BM25)& $7.6$ & $15.9$ & $28.8$ & $38.5$ & $34.3$ & $50.8$ \\ 
IRCoT (ColBERTv2) & $12.2$ & $24.1$ & $32.4$ & $43.6$ & $45.2$ & $63.7$ \\
HippoRAG w$/$ IRCoT & $\underline{14.2}$ & $25.9$ & $\underline{45.6}$ & $\underline{59.0}$ & $\underline{49.2}$ & $\underline{67.9}$ \\ 
\gear & $\mathbf{19.0}$ & $\mathbf{35.6}$ & $\mathbf{47.4}$ & $\mathbf{62.3}$ & $\mathbf{50.4}$ & $\mathbf{69.4}$ \\ \bottomrule
\end{tabular}
\caption{\textbf{End-to-end QA performance} using the top-$5$ retrieved passages. The \textbf{best} model is in bold and \underline{second best} is underlined. The top part shows the lower and upper bounds of QA performance, while the middle and bottom sections display scores for single-step and multi-step retrievers, respectively.}
\label{tab:em_and_f1}
\end{table}

\section{Discussion}

\begin{figure*}[thbp]
  \centering
  \includegraphics[width=0.92\textwidth]{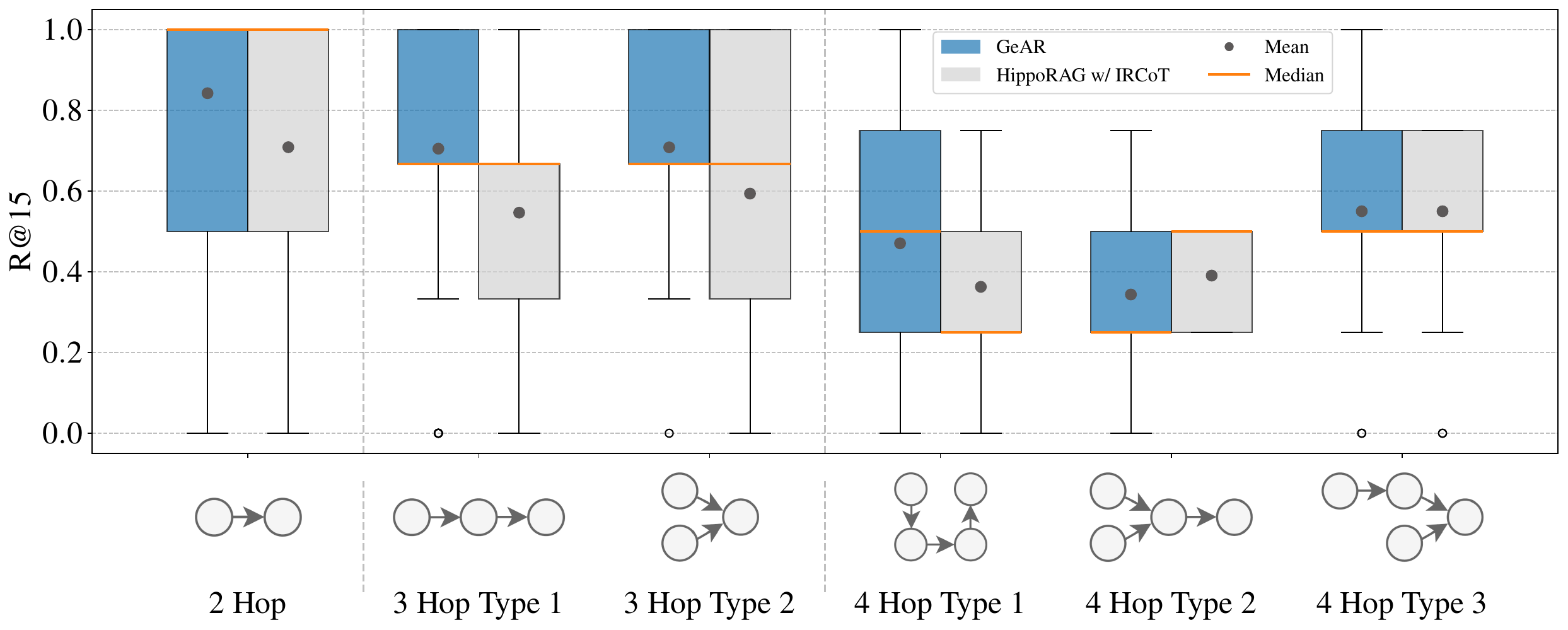}
  \caption{Analysis of R@15 performance divided by hop types on MuSiQue. The hop categorisation follows the MuSiQue documentation. Mean recall values are indicated by grey dots for each hop type.}
  \label{fig:recall_across_hoptype}
\end{figure*}

\subsection{What makes \gear work?}
\label{subsec:what_makes_gear_work}

\paragraph{NaiveGE vs SyncGE}
As shown in Table \ref{tab:recall_main_table}, both graph expansion variants enhance every base retriever's performance across all datasets. The superior performance of SyncGE indicates the effectiveness of using LLMs for locating initial nodes. Notably, it surpasses HippoRAG w$/$ IRCoT's on MuSiQue without multiple iterations.

\paragraph{Diverse Triple Beam Search improves performance}As shown in Table \ref{tab:diverse_beam_search}, our diverse triple beam search consistently outperforms standard beam search across all datasets and recall ranks. By incorporating diversity weights into beam search, we align a language modelling-oriented solution with information retrieval objectives that involve satisfying multiple information needs underlying multi-hop queries~\cite{Drosou2010}.

\begin{table}[t]
\small
\centering
\begin{tabular}{l l cc}
\toprule
\textbf{Metric} & \textbf{Dataset} & \textbf{w}$\mathbf{/}$ \textbf{Diversity} & \textbf{w}$\mathbf{/}$\textbf{o Diversity} \\ 
\midrule
\multirow{3}{*}{R@5} & MuSiQue & $\mathbf{48.7}$ & $47.0$ \\
& 2Wiki & $\mathbf{72.6}$ & $68.2$ \\
& HotpotQA & $\mathbf{87.4}$ & $85.0$ \\
\midrule
\multirow{3}{*}{R@10} & MuSiQue & $\mathbf{57.7}$ & $53.9$ \\
& 2Wiki & $\mathbf{80.9}$ & $76.0$ \\
& HotpotQA & $\mathbf{93.3}$ & $92.2$ \\
\midrule
\multirow{3}{*}{R@15} & MuSiQue & $\mathbf{61.2}$ & $58.4$ \\
& 2Wiki & $\mathbf{82.4}$ & $77.4$ \\
& HotpotQA & $\mathbf{95.2}$ & $94.3$ \\
\bottomrule
\end{tabular}
\caption{\textbf{Effects of beam search diversity} on Hybrid + SyncGE across MuSiQue, 2Wiki and HotpotQA.}
\label{tab:diverse_beam_search}
\end{table}

\paragraph{\gear \textrm{\textit{mostly}} nails it the first time}

While \gear supports multiple iterations, Figure~\ref{fig:recall_across_iterations} shows that \gear achieves strong retrieval performance in a single iteration on MuSiQue. This differentiates it from IRCoT-oriented setups that require at least $2$ iterations to reach maximum performance. This can be attributed to the fact that \gear reads (Eq.~\ref{eq:proximal_read_agent}) multi-hop contexts and associates proximal triples in gist memory with passages, establishing synergy between our graph retriever and the LLM. We believe this mirrors the hippocampal process of forming and resolving sparse representations, where gist memories are learned in a one or few-shot manner \cite{Hanslmayr2016}. The $10\%$ performance gap between Hybrid + SyncGE and \gear at $n=1$ indicates that the LLM reading and linking processes effectively approximate the hippocampus's role within our framework.

\subsection{How robust is \gear?}
\paragraph{\gear excels at questions of low-to-moderate complexity}

Figure \ref{fig:recall_across_hoptype} presents a detailed breakdown of retrieval performance across different hop types, including path-finding
and path-following questions categorized by \citeauthor{Gutierrez2024}. For 2-hop questions, while \gear and HippoRAG w$/$ IRCoT achieve similar interquartile ranges, \gear shows a higher mean recall, indicating superior performance on low-complexity questions. This advantage becomes more pronounced with 3-hop questions, where \gear's entire interquartile range exceeds HippoRAG w$/$ IRCoT's median performance across both hop subdivisions. This demonstrates \gear's enhanced capability in handling moderately complex questions. In addition to MuSiQue, 2Wiki and HotpotQA, we test \gear against the \textit{hand-picked} case study data provided by \citeauthor{Gutierrez2024}. These include four path-finding questions across four different domains. Our findings (Appendix~\ref{appendix_sec:hipporag_data}) indicate that \gear's performance is on par with HippoRAG w$/$ IRCoT, outperforming the competition in three out of the four cases, in terms of recall.

\paragraph{\gear's performance remains consistent across chunks with varying numbers of triples}
Using MuSiQue, we group questions based on the average number of triples (i.e. triple density, $\rho_t$) associated with their golden passages, and evaluate R@15 across four ranges: \begin{inparaenum}[(i)] \item $\rho_t < 9$, \item $9 \leq \rho_t < 11$, \item $11 \leq \rho_t < 13$ and \item $13 \leq \rho_t$\end{inparaenum}. Across all these ranges, the recall performance of both SyncGE and GeAR exhibits lower variation, with significantly smaller standard deviations of $1.18 \ll 2.04$ and $2.08 \ll 5.59$, respectively, compared to NaiveGE and HippoRAG w$/$ IRCoT. Further details are provided in Appendix~\ref{appendix_sec:robustness}.

\subsection{Is \gear efficient?}
\label{subsec:gear_efficient}
\begin{figure}[thbp]
  \includegraphics[width=\linewidth]{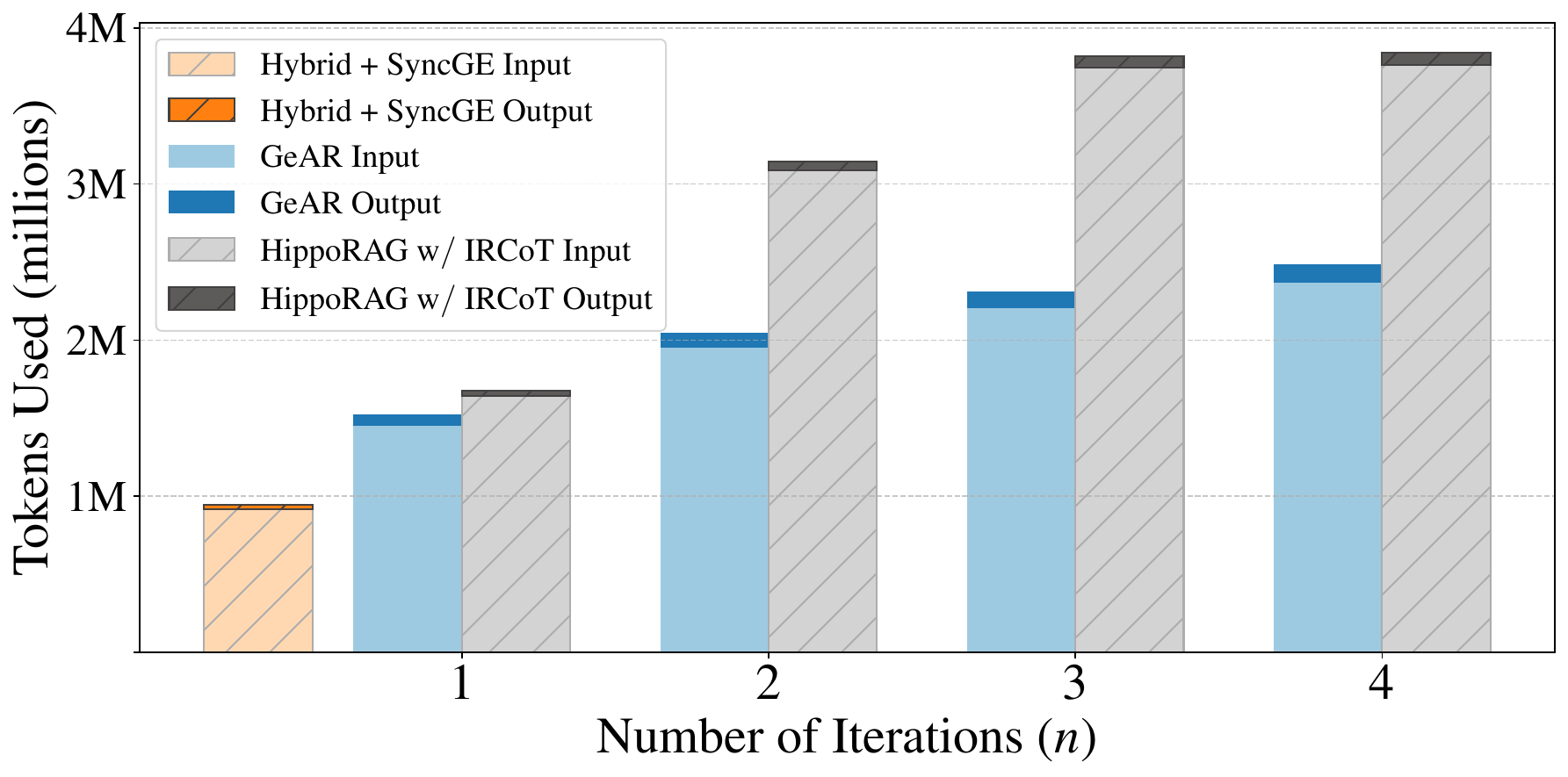}
  \caption{\textbf{Progressive accumulation of input and output LLM tokens across agent iterations} on MuSiQue.
  The Hybrid + SyncGE method appears only to the left of Iteration $1$ as it is a single-step approach.}
\label{fig:token_comparison_input_output}
\end{figure}

As observed in Figure~\ref{fig:recall_across_iterations}, \gear requires fewer iterations than the competition to reach its maximum recall performance. 
Furthermore, Figure~\ref{fig:token_comparison_input_output} shows that \gear can act as a more efficient alternative with respect to LLM token utilisation
.
We note that even for a single iteration, \gear uses fewer tokens than HippoRAG w$/$ IRCoT. In contrast to ours, this trend exacerbates for the competition as the number of iterations increases. These findings also reiterate the value of SyncGE, which outperforms a significantly more LLM-heavy solution on MuSiQue, using almost $2.9$ million fewer tokens. Even in the case that HippoRAG w$/$ IRCoT runs for a single iteration it would require more than $0.7$ million tokens that Hybrid + SyncGE, with a substantially lower R@15 of $51.7$.

\section{Conclusion}
\label{sec: conclusion}
We propose \gear, a novel framework that incorporates a graph-based retriever within a multi-step retrieval agent to model the information-seeking process for multi-hop question answering. 

We showcase the synergy between our proposed graph retriever (i.e. SyncGE) and the LLM within the \gear framework. SyncGE leverages the LLM to synchronise information from passages with triples and expands the graph by exploring diverse beams of triples that link multi-hop contexts. Our experiments reveal that this strategy improves over more naive implementations, demonstrating the LLM's capability to guide the exploration of initial nodes for graph expansion. Furthermore, \gear utilises multi-hop contexts returned by SyncGE and constructs a gist memory which is used for effectively summarising information across iterations. \gear achieves superior performance compared to other multi-step retrieval methods while requiring fewer iterations and LLM tokens.

\section*{Limitations}
\label{sec:limitations}
The scope of this paper is limited to retrieval using a graph of triples that bridge corresponding passages. While we demonstrate the effectiveness of our graph expansion approach and \gear{}, several components could be further refined. More sophisticated graph construction methods addressing challenges such as entity disambiguation \cite{Dredze2010} and knowledge graph completion \cite{Lin2015} may yield further improvements, as discussed in Appendix \ref{appendix_sec:graph_construction}. Similarly, our choice of a dense embedding model for the scoring function in the diverse triple beam search could be replaced by alternatives, such as formulating this as a natural language inference task \cite{Wang2021}.

Beyond the core graph and scoring mechanisms, our design choices for other agent components—such as the gist memory, reasoner, and query rewriter—adopt commonly validated practices from prior research \cite{Trivedi2022, Li2024, Fang2024}. While effective, exploring more sophisticated designs specifically tailored for these functions presents an opportunity for future performance gains.

\bibliography{main}

\appendix

\clearpage
\section{Dataset Choices and Statistics}
\label{appendix:dataset_stats}

\begin{table}[ht]
\centering
\small
\begin{tabular}{@{}lrrr@{}}
\toprule
& \textbf{MuSiQue} & \textbf{2Wiki} & \textbf{HotpotQA} \\ \midrule
Split Source & IRCoT & IRCoT & HippoRAG \\ \midrule
\# Hops  & $2-4$ & $2$ & $2$ \\
\# Documents & $139,416$ & $430,225$ & $9,221$ \\
\# Test Queries & $500$ & $500$ & $1,000$\\ \midrule
\# Chunks ($\mathbf{C}$) & $148,793$  & $490,454$ & $10,293$ \\
\# Triples ($\mathbf{T}$) & $1,521,136$  & $4,993,637$ & $122,492$ \\
Av. \# $\mathbf{T}/\mathbf{C}$ & $10.2$  & $10.2$ & $11.9$ \\
\bottomrule
\end{tabular}
\caption{Dataset characteristics and preprocessing statistics, where triples are extracted from chunks, and Av. \# $\mathbf{T}$$/$$\mathbf{C}$ represents the average number of triples per chunk.}
\label{tab:dataset statistics}
\end{table}

Table \ref{tab:dataset statistics} serves as a summary of various facts and statistics related to the employed datasets and the chunking and triple extraction process introduced in Section \ref{sec:preliminaries}. Please note for all the evaluated datasets, we use their open-domain setting and answerable subset if applicable.

\paragraph{Reasoning behind dataset split choices}
For MuSiQue and 2Wiki, we use the data provided by \citeauthor{Trivedi2023}, including the full corpus and sub-sampled test cases for each dataset. To limit the experimental cost for HotpotQA, we follow the setting by \citeauthor{Gutierrez2024} where both the corpus and test split are smaller than IRCoT's counterpart.
\section{More Implementation Details}
\label{appendix_sec:detailed_implementation_details}

\subsection{Baselines Details}
\label{appendix:retrievers_implementation_details}
We implement all proposed approaches using Elasticsearch\footnote{\url{https://www.elastic.co}}. For SBERT, we employ the \texttt{all-mpnet-base-v2} model with approximate k-nearest neighbours and cosine similarity for vector comparisons. In IRCoT experiments, we evaluate both ColBERTv2 and BM25 retrievers — ColBERTv2 for alignment with HippoRAG's baselines, and BM25 for consistency with the original IRCoT implementation.

For all multi-step approaches, including ours, we follow \citeauthor{Gutierrez2024} with respect to the maximum number of retrieval iterations, which vary based on the hop requirements of each dataset. Thus, we use a maximum of 4 iterations for MuSiQue and 2 iterations for HotpotQA and 2Wiki.

\subsection{\gear Details}
\gear involves several hyperparameters, such as the beam size inside graph expansion. 
We randomly sampled $500$ questions from the MuSiQue development set, which we ensure not to overlap with the relevant test set. We select our hyperparameters based on this sample without performing a grid search across all possible configurations. Our goal is to demonstrate that our method is able to achieve state-of-the-art results without extensive parameter tuning. 


The initial retrieval phase utilises the chunks index $\mathbf{C}$ as the information source, while leaving the triple index $\mathbf{T}$ unused. Our graph expansion component implements beam search with length 2, width 10, and 100 neighbours per beam. The hyperparameter $\gamma$ employed in diverse triple beam search is set to twice the beam search width. For the scoring function, we use the cosine similarity score and the SBERT embedding model. In Appendix~\ref{appendix_sec:ablation_studies}, we test the performance of \gear across different beam search length values and maximum numbers of agent iterations.

For the single-step configurations (i.e. any base retriever with NaiveGE or SyncGE), we set the base retriever's maximum number of returned chunks to match our evaluation recall threshold. With the multi-step setup, we maintain a consistent maximum of 10 retrieved chunks before knowledge synchronisation for the purpose of matching IRCoT's implementation. While this 10-chunk limitation applies to individual retrieval rounds, please note that the total number of accessible chunks can exceed this threshold through graph expansion and multiple \gear iterations.

\paragraph{\textrm{\texttt{passageLink}} Details\label{appendixpara:passage_link}}
We use \texttt{passageLink} to link each triple $t_j \in \mathcal{G}^{(n)}$ to its corresponding passages in $\mathbf{C}$ by running a retrieval step as follows:
\begin{align}
h^k_{\text{base}}\left( \mathbf{q}, {\mathbf{C} \cup \mathbf{T}} \right) 
    &= \text{RRF} \Big ( h^k_{\text{base}}\left( t_j, {\mathbf{C}}\right), \nonumber \\
    &\quad\quad h^k_{\text{base}}\left( t_j, {\mathbf{T}} \right) \Big ),
\end{align}
where $j \in \left \{1, \dots, \vert\mathcal{G}^{(n)}\vert \right \}$ and $h^k_{\text{base}}\left( t_j, {\mathbf{C} \cup \mathbf{T}} \right)$ is the RRF of passages returned by both $\mathbf{T}$ and $\mathbf{C}$ when queried with $t_j$. 

\section{Walkthrough Example}
\label{appendix_sec:walkthrough_example}

Table \ref{tab:walkthrough_example} provides a visual step-by-step example of our framework. The example demonstrates how each component contributes to the final retrieved passages, which we hope offers useful context for understanding our design decisions.

\begin{table*}[p]
\centering
\resizebox{\textwidth}{!}{
\begin{tabular}{@{} L{7cm} L{13cm} C{2.5cm} @{}} 

\toprule 

\textbf{Module} & \textbf{Intermediate Output} & \textbf{Fig.\ref{fig:system_diagram} Sections}  \\

\midrule

\textbf{\underline{\smash{Offline Index Building Stage}}}\newline\vspace{0.2cm}
For each passage $c_i \in \mathbf{C} = \left \{c_1, c_2, \ldots, c_C \right \}$, an LLM extracts a triple set, such that each triple is uniquely linked to one single passage.
&
\adjustimage{width=0.65\textwidth,valign=b}{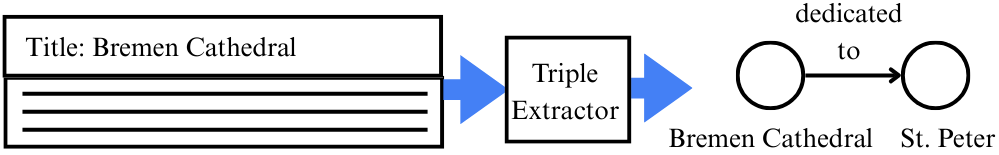} 
& \\

\midrule
\midrule

\multicolumn{3}{@{}l}{$\downarrow$~\textbf{\underline{\smash{Input Query}}} from MuSiQue~\cite{Trivedi2022} \qquad (\textbf{\underline{\smash{Answer}}}: 1929)} \\

\multicolumn{2}{@{}l}{When did the location of the basilica which is named for the same saint that the Bremen Cathedral is named for become a country?}

&\\

\midrule
\midrule

\multicolumn{3}{@{}l}{\textbf{\underline{\smash{Online Retrieval Stage}}}} \\
\addlinespace 

\textbf{1. Base Retrieval} (see~\S\ref{sec:graph_retrieval}) \newline
For a query $\mathbf{q}$, $\mathbf{C}_\mathbf{q}' = h^k_{\text{base}}\left( \mathbf{q}, {\mathbf{C}}\right )$  is a list of passages given by the retriever, implemented as BM25, SBERT, or a mix of both.
&
$\bm{\mathcal{P}_{1}}$ Bremen Cathedral, \quad 
$\bm{\mathcal{P}_{2}}$ Münster Cathedral, \quad 
$\bm{\mathcal{P}_{3}}$ Basilica of the Sacred Heart \newline
$\bm{\mathcal{P}_{4}}$ Saint Justin's Church, Frankfurt-Höchst, \qquad~~\,
$\bm{\mathcal{P}_{5}}$ Alatri Cathedral
&

\adjustimage{width=0.15\textwidth,center}{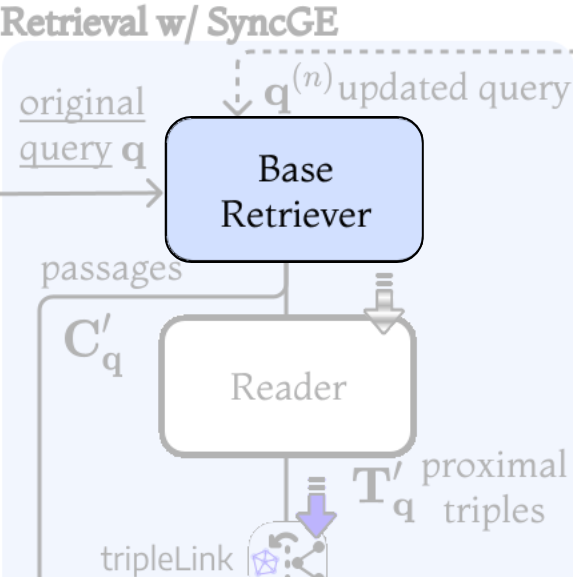} \\

\midrule
\textbf{2. Reader} (see~\S\ref{subsection:knowledge_syncro}) \newline\vspace{0.2cm}
An LLM \texttt{reads} $\mathbf{C}_\mathbf{q}'$  and summarises knowledge triples, outputting a collection $\mathbf{T}_\mathbf{q}'$ of triples: the \textit{proximal triples}.
&
$\bm{\mathcal{T}_{1}'}$ $\langle$Bremen Cathedral, dedicated to, St.\,Peter$\rangle$ \newline
$\bm{\mathcal{T}_{2}'}$  $\langle$Alatri Cathedral, dedicated to, Saint Paul$\rangle$ \newline
$\bm{\mathcal{T}_{3}'}$  $\langle$Alatri Cathedral, co-cathedral of, Diocese Anagni-Alatri$\rangle$ \newline
$\bm{\mathcal{T}_{4}'}$  $\langle$Bremen, is located in, Germany$\rangle$
&
\adjustimage{width=0.15\textwidth,center}{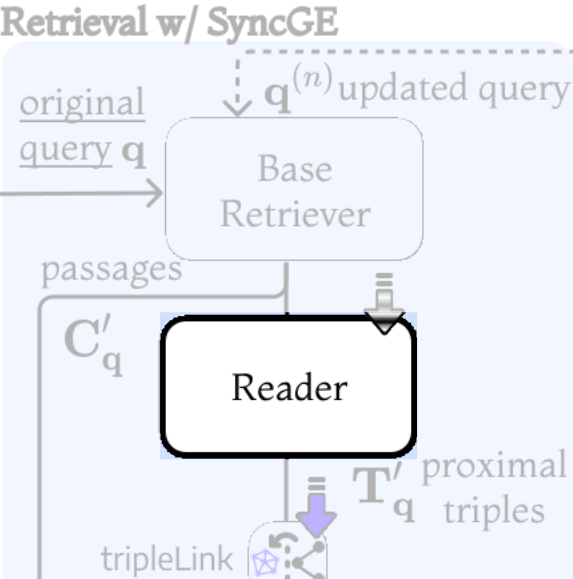} \\

\midrule
\textbf{2. \linkTriple{}} (see~\S\ref{subsection:knowledge_syncro})\newline\vspace{0.2cm}
Initial nodes $\Tqinit$ for graph expansion are identified by linking each triple in $\mathbf{T}_\mathbf{q}'$ to a triple in $\mathbf{T}$, using the \linkTriple{} function.
&
$\bm{\mathcal{T}_{1}}$  $\langle$Bremen Cathedral, dedicated to, St.\,Peter$\rangle$ \newline
$\bm{\mathcal{T}_{2}}$  $\langle$Alatri Cathedral, dedicated to, Saint Paul$\rangle$ \newline
$\bm{\mathcal{T}_{3}}$ $\langle$Diocese of Macerata-Tolentino-Recanati-Cingoli-Treia, type, co-cathedral$\rangle$ \newline
$\bm{\mathcal{T}_{4}}$ $\langle$Bremen, part of, Germany$\rangle$
&
\adjustimage{width=0.15\textwidth,center}{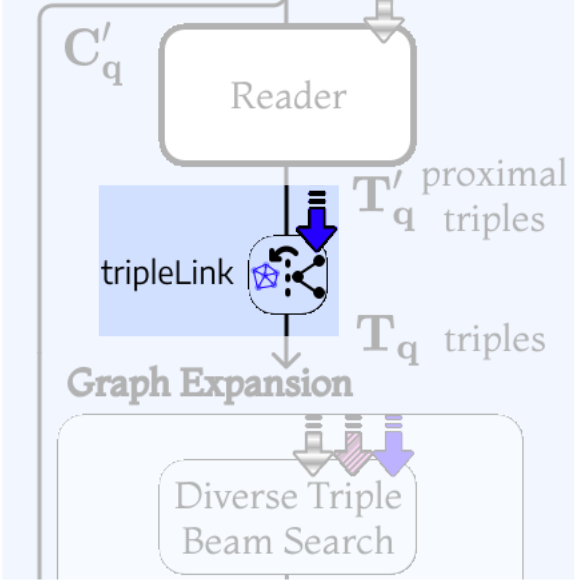} \\

\midrule
\textbf{4. Graph Expansion} (see~\S\ref{subsec:diverse_triple_beam_search})\newline\vspace{0.2cm}
The primary component of graph expansion is \textit{Diverse Triple Beam Search}. Here, we explore
neighbourhood of a triple (defined as other triples with shared head or tail entities) and
maintain top-$b$ sequences (beams) of triples.
&
\adjustimage{width=4.3cm,valign=b,clip=true,trim=0.1cm 0cm 0cm 0.2cm}{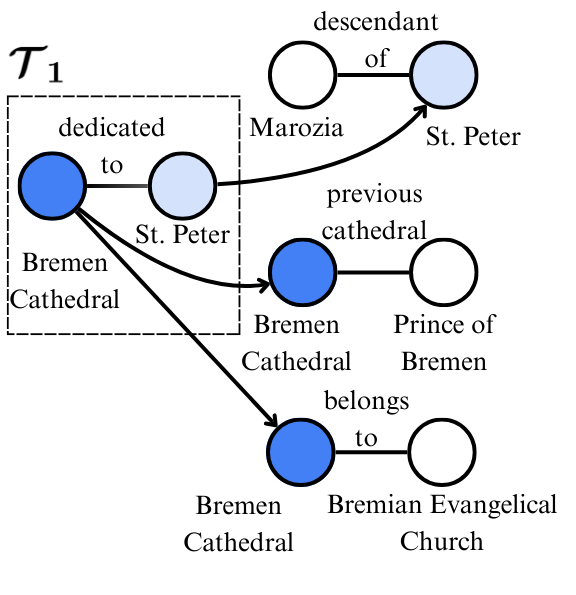}\hfill
\adjustimage{width=4.3cm,valign=b,clip=true,trim=0.1cm 0cm 0cm 0.2cm}{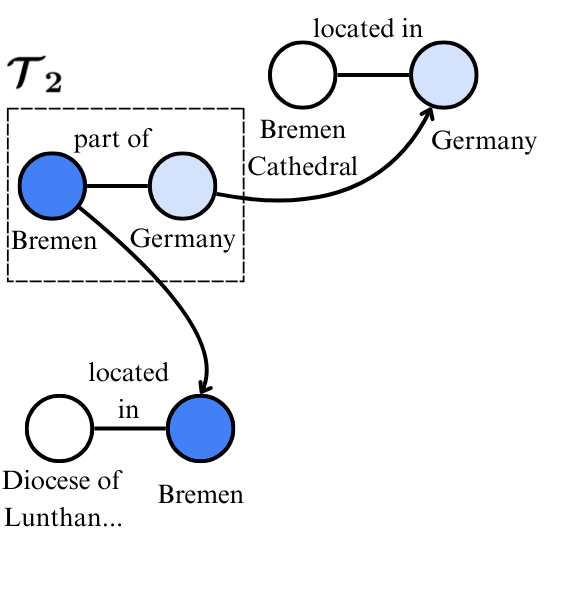}\hfill
\adjustimage{width=4.3cm,valign=b,clip=true,trim=0.1cm 0cm 0cm 0.2cm}{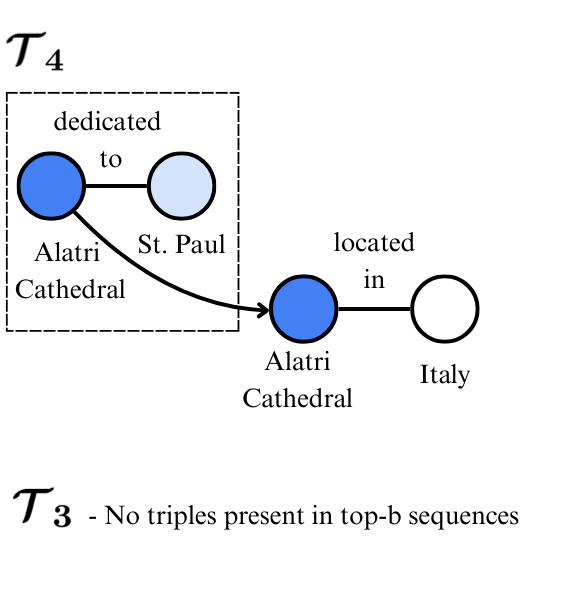}
&
\adjustimage{width=0.15\textwidth,center}{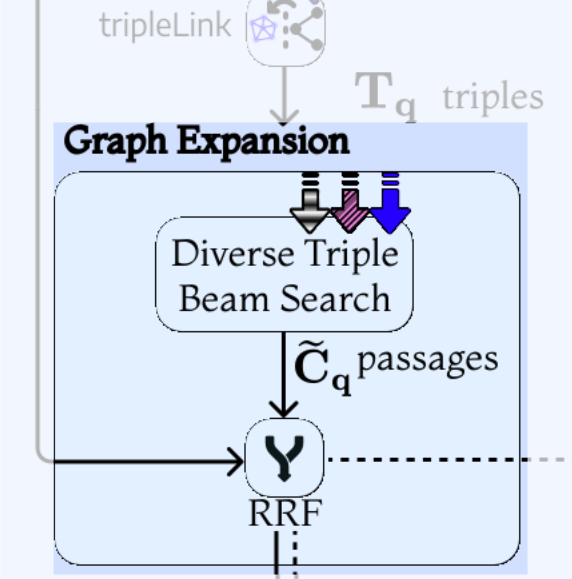} \\

\midrule
\textbf{5. Gist Memory} (see~\S\ref{subsec:gist_memory_constructor})\newline\vspace{0.2cm}
Similar to the Reader, an LLM reads a collection of retrieved paragraphs $\mathbf{C}_{\mathbf{q}^{(n)}}$ and extracts an array of proximal triples $\mathbf{T}_{\mathbf{q}^{(n)}}^{\mathcal{G}}$, which are stored in the Gist Memory $\mathcal{G}^{(n)}$.
&
$\bm{\mathcal{T}_{1}}$  $\langle$Bremen Cathedral, dedicated to, St.\,Peter$\rangle$ \newline
$\bm{\mathcal{T}_{2}}$  $\langle$Alatri Cathedral, dedicated to, Saint Paul$\rangle$ \newline
$\bm{\mathcal{T}_{3}}$ $\langle$Lund Cathedral, dedicated to, Saint Lawrence$\rangle$ \newline
$\bm{\mathcal{T}_{4}}$ $\langle$Bremen, part of, Germany$\rangle$
&
\adjustimage{width=0.15\textwidth,center}{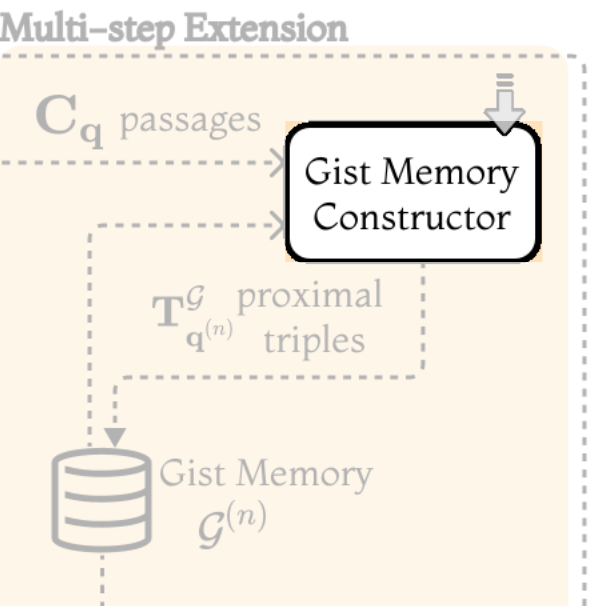} \\

\midrule
\textbf{6. Reasoner} (see~\S\ref{subsec:reasoning})\newline\vspace{0.2cm}
After updating $\mathcal{G}^{(n)}$, we assess whether it contains sufficient evidence to answer the original question via an LLM reasoning step.
&
\textbf{Answerable:} False \qquad\qquad
\textbf{Answer or reason:} The provided facts do not contain information about the location of the basilica named for St.\,Peter, nor do they provide any details about when it became a country. The facts only mention the dedication of other cathedrals to different saints.
&
\adjustimage{width=0.15\textwidth,center}{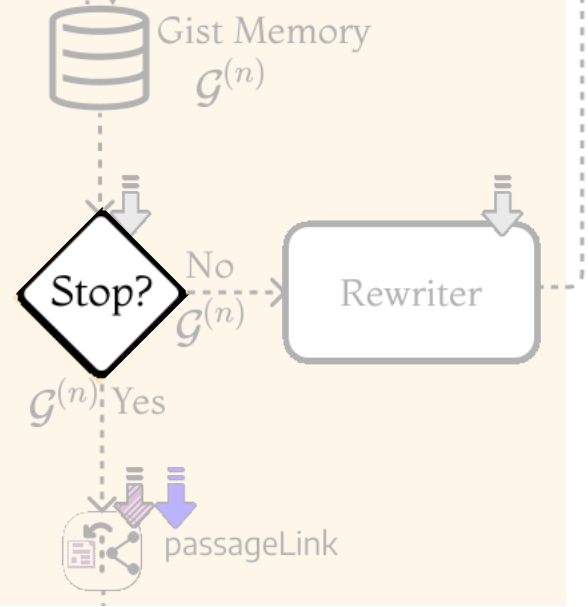} \\

\midrule
\textbf{7. Rewriter} (see~\S\ref{subsec:rewriting})\newline\vspace{0.2cm}
Given the original $\mathbf{q}$, the accumulated memory $\mathcal{G}^{(n)}$, and the reasoning output $\mathbf{r}^{(n)}$, an LLM is used to re-write the query.\newline We return to \textbf{step 1} and repeat.
&
\textbf{Next query:} What is the location of the basilica dedicated to St. Peter, and when did that location become a country?
&
\adjustimage{width=0.15\textwidth,center}{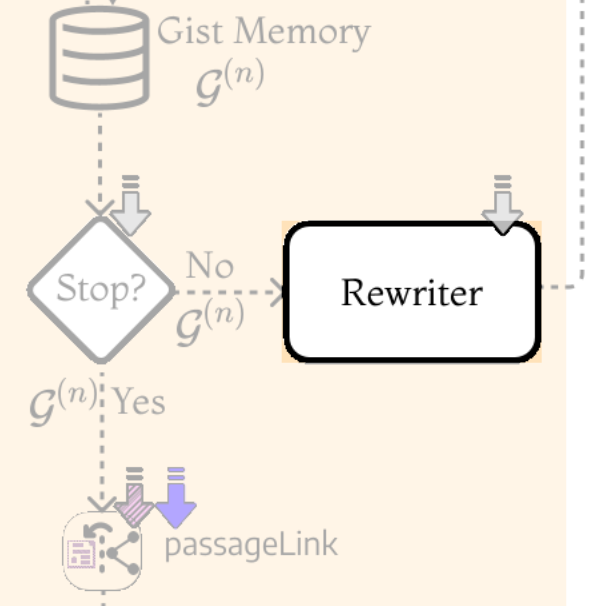} \\

\bottomrule 

\end{tabular}
}

\caption{\textbf{Visual walk-through example} of the modules involved in offline index construction and online retrieval in \gear. After query-rewriting, steps 1-7 are repeated until termination by the Reasoner, or reaching the maximum number of iterations. }
\label{tab:walkthrough_example} 

\end{table*}

\section{Ensuring Fair Comparisons}
\label{appendix_sec:hipporag_results_original_prompt}

\begin{table*}[t]
\small
\centering
\begin{tabular}{l l ccc ccc ccc}
\toprule
& & \multicolumn{3}{c}{\textbf{MuSiQue}} & \multicolumn{3}{c}{\textbf{2Wiki}} & \multicolumn{3}{c}{\textbf{HotpotQA}} \\ 
\cmidrule(lr){3-5} \cmidrule(lr){6-8} \cmidrule(lr){9-11}
& & R@5 & R@10 & R@15 & R@5 & R@10 & R@15 & R@5 & R@10 & R@15 \\ 
\midrule
\multirow{2}{*}{\parbox[t]{2.5cm}{\textbf{HippoRAG}}}
& original prompt & \textbf{41.9} & 46.9 & 51.1 & \textbf{75.4} & \textbf{83.5} & \textbf{86.9} & 79.7 & 88.4 & 91.4 \\ 
& our prompt & 41.0 & \textbf{47.0} & \textbf{51.4} & 75.1 & 83.2 & 86.4 & \textbf{79.8} & \textbf{89.0} & \textbf{92.4} \\ 
\midrule
\multirow{2}{*}{\parbox[t]{2.5cm}{\textbf{HippoRAG \\ w$/$ IRCoT}}}
& original prompt & \textbf{49.9} & \textbf{56.4} & \textbf{59.3} & 81.5 & 90.2 & 92.3 & \textbf{90.2} & \textbf{94.7} & 95.8 \\ 
& our prompt & 48.8 & 54.5 & 58.9 & \textbf{82.9} & \textbf{90.6} & \textbf{93.0} & 90.1 & \textbf{94.7} & \textbf{95.9} \\ 
\bottomrule
\end{tabular}
\caption{Retrieval performance comparison between HippoRAG's sequential triple extraction method and our joint extraction approach across three datasets.}
\label{tab:hippo_prompt_vs_our_prompt}
\end{table*}

\begin{table*}[t]
\small
\centering
\begin{tabular}{@{}llcccccccccc@{}}
\toprule
\multirow{2}{*}{\textbf{LLM}} & \multirow{2}{*}{\textbf{Retriever}} & \multicolumn{3}{c}{\textbf{MuSiQue}} & \multicolumn{3}{c}{\textbf{2Wiki}} & \multicolumn{3}{c}{\textbf{HotpotQA}}\\
\cmidrule(lr){3-5} \cmidrule(lr){6-8} \cmidrule(lr){9-11}
& & R@5 & R@10 & R@15 & R@5 & R@10 & R@15 & R@5 & R@10 & R@15\\ \midrule
\multirow{3}{*}{\textbf{GPT-3.5 turbo}} & Hybrid + SyncGE & 48.3 & 55.0 & 58.1 & 70.7 & 78.4 & 79.7 & 86.1 & 92.0 & 94.3 \\
& HippoRAG w$/$ IRCoT & 45.5 & 51.0 & 54.8 & 80.0 & 87.9 & 89.8 & 87.4 & 92.6 & 94.6 \\
& \gear & \textbf{52.9} & \textbf{62.1} & \textbf{64.4} & \textbf{84.2} & \textbf{90.0} & \textbf{90.3} & \textbf{91.0} & \textbf{95.6} & \textbf{96.1} \\ \bottomrule
\end{tabular}
\caption{Retrieval performance for our proposed retrievers and HippoRAG w/ IRCoT across various datasets. We use \texttt{gpt-3.5-turbo-1106} (temperature $=$ 0) as the underlying LLM, to replicate HippoRAG's experimental setup.}
\label{tab:using_gpt3.5turbo}
\end{table*}

Although related studies often use common datasets, their experimental settings are frequently inconsistent. For instance, \citeauthor{Gutierrez2024} used heavily sub-sampled corpora, drastically reducing the number of documents (e.g., from over 5M to ~9k for HotpotQA) compared to the full datasets originally established for IRCoT \citep{Trivedi2022}. We believe such reductions significantly simplify the retrieval task.
Therefore, in our paper, we reproduced HippoRAG on MuSiQue and 2Wiki using the same dataset settings (i.e., full corpus and identical evaluation split) as defined in the original IRCoT paper \citep{Trivedi2022}. On HotpotQA, however, processing the full corpus established by \citet{Trivedi2022} (over 5 million passages) was computationally prohibitive, so we follow the same setting as HippoRAG to limit the experimental cost.

To ensure fairness in our comparisons, we ran all baselines using a consistent experimental setup. Additionally, for areas of potential discrepancy, and where possible, we report the retrieval performance of baselines in their original configuration to confirm that the used experimental setup does not adversely affect performance. Below, we address any potential discrepancies in the following areas: \begin{inparaenum}[(i)] \item triple extraction methodology, \item retrieval metrics, and \item LLMs\end{inparaenum}.

\paragraph{Choice of triple extraction methodology}
HippoRAG employs a sequential approach to triple extraction: it first identifies named entities from a text chunk, and then uses these entities to guide triple extraction in a second step. In contrast, our method extracts both entities and triples simultaneously. Table \ref{tab:hippo_prompt_vs_our_prompt} shows that both approaches achieve comparable retrieval performance across all datasets, with each method excelling in different scenarios. These results validate that joint entity and triple extraction can match the effectiveness of sequential extraction while reducing the number of required processing steps.

\paragraph{Reasoning behind retrieval metrics}
\label{appendix:reasoning_behind_retrieval_metrics}
The MuSiQue dataset contains 2-hop, 3-hop and 4-hop questions, where a $k$-hop question is defined as one that requires $k$ pieces of evidence to reach the correct answer \cite{Trivedi2022}. This means 3-hop and 4-hop questions require more than 2 pieces of evidence. If we used Recall@2 to evaluate them, as previous works such as \cite{Gutierrez2024} do, we would be misjudging these questions, since it assumes only two pieces of evidence are enough for perfect recall. Additionally, given modern LLMs' expanding context length capabilities \cite{Ding2024}, examining recall beyond R@5 (HippoRAG's highest evaluated rank) provides valuable insights. Following IRCoT's approach, we measure up to R@15 and include R@10 as an intermediate point, offering a comprehensive view of model performance across retrieval depths. Therefore, our evaluation employs recall at ranks 5, 10, and 15 (R@5, R@10, R@15).

\paragraph{Choice of LLM}
\label{appendix_para:llm_choice}
\citeauthor{Gutierrez2024} use \texttt{gpt-3.5-turbo-1106} for their experiments, whereas in this paper we reproduce it with GPT-4o mini. GPT-4o mini was selected as a more capable alternative to GPT-3.5 Turbo (please refer to: \href{https://openai.com/index/gpt-4o-mini-advancing-cost-efficient-intelligence}{\texttt{https://openai.com/index/gpt-4\allowbreak -mini-advancing-cost-efficient-\allowbreak intelli\-gence}}). In order to alleviate any concerns regarding discrepancies with respect to the selected LLM, we also run experiments using \texttt{gpt-3.5-turbo-1106}. Table \ref{tab:using_gpt3.5turbo} shows the retrieval results of our proposed methods against HippoRAG w$/$ IRCoT. We observe a similar trend to that in Table \ref{tab:recall_main_table}---\gear surpasses the performance of HippoRAG w$/$ IRCoT.


\section{Why this graph construction method?}
\label{appendix_sec:graph_construction}

We adopt an LLM-based triple extraction methodology, following the approach outlined in HippoRAG \cite{Gutierrez2024}. In their study, they evaluated the performance of various LLMs in OpenIE and compared these results with those of the end-to-end REBEL model \cite{HuguetCabot2021}. They reported substantial improvements in triple extraction when using LLMs in domains that deviate from conventional ClosedIE or OpenIE settings, which are respectively overly constrained and unconstrained in terms of named entities and pre-defined relations. Similar concerns about the generalisability and scalability of conventional KG construction approaches in open-domain scenarios are recognised by \citeauthor{Wang2024}, who sought to construct their graphs without relying on pre-existing ontologies, or KGs for named entity disambiguation.

These findings resonate with the growing interest in recent literature towards applying such methodologies for automatic, schema-free knowledge graph construction \cite{Li2024, Fang2024, Gutierrez2024, Park2024}. As our primary focus is retrieval rather than graph construction, we adopt the triple extraction methodology from HippoRAG and refer readers to their paper for a more detailed analysis.

Our work presents a novel framework for advancing the performance of RAG systems in the context of texts associated with schema-free triples. We follow HippoRAG's graph construction approach, as exploring graph construction methods falls outside the scope of this paper. However, this does not imply that our proposed method relies on this specific approach, and we believe that further improvements in graph construction could lead to additional gains.

\section{\gear across Different Configurations}
\label{appendix_sec:ablation_studies}

\begin{table*}[t]
\small
\centering
\resizebox{\textwidth}{!}{%
\begin{tabular}{l l ccc ccc ccc ccc}
\toprule
& & \multicolumn{12}{c}{\textbf{R@$k$ across Different Maximum Numbers of Iterations}} \\ 
\cmidrule(lr){3-14}
& & \multicolumn{3}{c}{$n$ = 1} & \multicolumn{3}{c}{$n$ = 2} & \multicolumn{3}{c}{$n$ = 3} & \multicolumn{3}{c}{$n$ = 4} \\ 
\cmidrule(lr){3-5} \cmidrule(lr){6-8} \cmidrule(lr){9-11} \cmidrule(lr){12-14}
& & R@5 & R@10 & R@15 & R@5 & R@10 & R@15 & R@5 & R@10 & R@15 & R@5 & R@10 & R@15 \\ 
\midrule
\multirow{3}{*}{\parbox{2cm}{\textbf{Beam Search\\Length}}}
& $b$ = 2 & 58.1 & 66.0 & 69.5 & \textbf{59.2} &\textbf{ 68.9 }& 71.3 & 57.9 & 68.0 & \textbf{71.5} & 58.4 & 67.6 & \textbf{71.5} \\
& $b$ = 3 & 55.9 & 64.6 & 67.9 & 57.2 & 66.6 & 70.2 & 58.1 & 67.8 & 71.0 & 56.7 & 66.1 & 70.4 \\
& $b$ = 4 & 54.9 & 62.9 & 67.3 & 56.6 & 66.3 & 69.3 & 58.1 & 67.9 & 71.0 & 56.1 & 66.1 & 69.9 \\
\bottomrule
\end{tabular}}
\caption{\gear's retrieval performance across different hyper-parameters in terms of maximum number of agent iterations ($n$) and graph expansion's beam search length ($b$). Results are reported using Recall@$k$ (R@$k$) for $k \in \left \{5, 10, 15 \right\}$ for the MuSiQue dataset.}
\label{tab:varying_beam_length_and_agent_iterations}
\end{table*}

Table \ref{tab:varying_beam_length_and_agent_iterations} illustrates the performance of \gear across varying hyperparameter configurations, including beam search length—applied during graph expansion—and the maximum number of agent iterations. As the maximum number of iterations increases, \gear achieves better retrieval performance. However, consistently with the trends shown in Figure \ref{fig:recall_across_iterations}, this improvement levels off when setting the maximum number of iterations at $n \geq 3$. In contrast, increasing beam sffearch length above $2$ slightly reduces performance.
Despite this, \gear maintains highly competitive results and significantly outperforms alternative methods shown in Table \ref{tab:recall_main_table}.

\section{Compatibility with Open-weight Models}
\label{appendix_sec:open_source_model_experiments}

\paragraph{\gear Results}
As shown in Table \ref{tab:open_source_recall}, we evaluate \gear using popular 7-8B parameter open-weight models, comparing them against a closed-source alternative. On HotpotQA, Llama-3.1-7B surpasses the closed-source alternative, achieving higher recall rates at R@10 and R@15. For MuSiQue and 2Wiki, while the closed-source model maintains a slight superior edge in performance, the margin is narrow. Importantly, all tested open-weight models consistently outperform the previous state-of-the-art, HippoRAG w$/$IRCoT. This decouples \gear from the need to use closed-source models, suggesting that state-of-the-art multi-step retrieval can be achieved using more accessible models.

\begin{table*}[p]
\small
\centering
\resizebox{\textwidth}{!}{%
\begin{tabular}{l L{2cm} ccc ccc ccc}
\toprule
& \multirow{2.5}{*}{\textbf{LLM}} & \multicolumn{3}{c}{\textbf{MuSiQue}} & \multicolumn{3}{c}{\textbf{2Wiki}} & \multicolumn{3}{c}{\textbf{HotpotQA}} \\ 
\cmidrule{3-11}
& & R@5 & R@10 & R@15 & R@5 & R@10 & R@15 & R@5 & R@10 & R@15 \\ 
\midrule
\multirow{1}{*}{\textbf{Closed-source}} 
& GPT-4o mini & $\mathbf{58.4}$ & $\mathbf{67.6}$ & $\mathbf{71.5}$ & $\mathbf{89.1}$ & $\mathbf{95.3}$ & $\mathbf{95.9}$ & $\mathbf{93.4}$ & $96.8$ & $97.3$ \\ 
\midrule
\multirow{2}{*}{\textbf{Open-weight}}
& Llama-3.1-8B & $52.4$ & $62.3$ & $66.7$ & $81.6$ & $91.0$ & $93.7$ & $92.2$ & $\textbf{97.4}$ & $\textbf{98.1}$ \\ 
& Qwen-2.5-8B & $53.7$ & $63.7$ & $66.7$ & $85.9$ & $91.6$ & $93.0$ & $91.7$ & $96.2$ & $96.9$ \\ 
\bottomrule
\end{tabular}
}
\caption{Retrieval performance of \gear across different closed-source and open-weight models on MuSiQue, 2Wiki and HotpotQA. Results are reported using Recall@$k$ (R@$k$) for $k \in \left \{5, 10, 15 \right\}$, showing the percentage of questions for which the correct entries are found within the top-$k$ retrieved passages. }
\label{tab:open_source_recall}
\end{table*}

\begin{table*}[p]
\small
\centering
\resizebox{\textwidth}{!}{%
\begin{tabular}{l l ccc ccc ccc}
\toprule
& & \multicolumn{3}{c}{\textbf{MuSiQue}} & \multicolumn{3}{c}{\textbf{2Wiki}} & \multicolumn{3}{c}{\textbf{HotpotQA}} \\ 
\cmidrule(lr){3-5} \cmidrule(lr){6-8} \cmidrule(lr){9-11}
& & R@5 & R@10 & R@15 & R@5 & R@10 & R@15 & R@5 & R@10 & R@15 \\ 
\midrule
\multirow{2}{*}{\textbf{GPT-4o mini}}
& w$/$ diversity & \textbf{48.7} & \textbf{57.7} & \textbf{61.2} & \textbf{72.6} & \textbf{80.9} & \textbf{82.4} & \textbf{87.4} & \textbf{93.3} & \textbf{95.2} \\ 
& w$/$o diversity & 47.0 & 53.9 & 58.4 & 68.2 & 76.0 & 77.4 & 85.0 & 92.2 & 94.3 \\ 
\midrule
\multirow{2}{*}{\textbf{Llama-3.1-8B-Instruct}}
& w$/$ diversity & $\textbf{46.2}$ & $\textbf{54.3}$ & $\textbf{57.4}$ & $\textbf{69.1}$ & $\textbf{78.1}$ & $\textbf{81.6}$ & $\textbf{87.3}$ & $\textbf{92.8}$ & $\textbf{95.1}$ \\ 
& w$/$o diversity & $44.9$ & $52.7$ & $55.0$ & $66.9$ & $75.9$ & $78.2$ & $85.0$ & $91.7$ & $94.4$ \\ 
\bottomrule
\end{tabular}}
\caption{Retrieval performance of the Hybrid + SyncGE method with different LLMs for the \texttt{read} step (see Eq.~\ref{eq:proximal_read}) w$/$ and w$/$o diversity for triple beam search. Results are reported using Recall@$k$ (R@$k$) for $k \in \left \{5, 10, 15 \right\}$, showing the percentage of questions for which the correct entries are found within the top-$k$ retrieved passages.}
\label{tab:diverse_beam_search_expanded}
\end{table*}

\paragraph{Diverse Beam Search Results}
\label{appendix_sec:diverse_beam_search_results_expanded}Expanding upon Table \ref{tab:diverse_beam_search}, Table \ref{tab:diverse_beam_search_expanded} demonstrates that diverse beam search consistently improves retrieval performance across both closed-source and open-weight models when using our proposed Hybrid + SyncGE setup. This further confirms the broader applicability of this approach.

\section{Robustness Studies}
\label{appendix_sec:robustness}
We assess the robustness of our framework in retrieving passages when triple extraction produces either limited or excessive triple content. Using the MuSiQue dataset, we group questions based on the average number of triples (i.e. triple density, $\rho_t$) associated with their golden passages and evaluate R@15 performance across these ranges. Table \ref{tab:robustness_table} presents the results for both the single- and multi-step retrieval settings. The passage length remains consistent across passages; hence, triple density serves as a proxy for the quality of triple extraction.

The results showcase that SyncGE and \gear are more robust than the competition at retrieving suitable passages. NaiveGE's performance tends to decline when the average number of triples associated with the gold passages either falls below or exceeds a certain threshold (for MuSiQue, the average number of triples extracted from the gold passages is $11.71$). A similar trend is observed for HippoRAG w$/$ IRCoT in the case of golden passages associated with more than 11 triples. We believe that this trend can be partially attributed to the Personalised PageRank machinery that makes HippoRAG agnostic to the semantic relationships of the extracted triples. In contrast, SyncGE and \gear are able to maintain consistent performance across both dense and sparse triple extraction outcomes.

\begin{table*}[p]
\small
\centering
\small
\begin{tabular}{@{}l@{\hspace{2pt}}lcccc@{}}
\toprule
& {\textbf{Retriever}} & $\rho_t < 9$ & $9 \leq \rho_t < 11$ & $11 \leq \rho_t <13$ & $13 \leq \rho_t$ \\ \midrule
\multirow{2}{*}{\parbox{3cm}{\textbf{Single-step Retrieval}}}
& Hybrid + NaiveGE & $50.6$ & $54.6$ & $54.1$ & $50.0$ \\
& Hybrid + SyncGE & $\mathbf{62.8}$ (↑ $12.2\%$) & $\mathbf{61.2}$ (↑ $6.6\%$) & $\mathbf{59.8}$ (↑ $5.7\%$) & $\mathbf{60.1}$ (↑ $10.1\%$) \\ 
\midrule
\multirow{2}{*}{\parbox{3cm}{\textbf{Multi-step Retrieval}}}
& HippoRAG w$/$ IRCoT & $64.4$ & $65.5$ & $55.0$ & $52.8$ \\
& \gear & $\mathbf{73.6}$ (↑ $9.2\%$) & $\mathbf{73.7}$ (↑ $8.2\%$) & $\mathbf{69.3}$ (↑ $14.3\%$) & $\mathbf{69.7}$ (↑ $16.9\%$) \\ \bottomrule
\end{tabular}
\caption{Retrieval performance for single- and multi-step retrievers across different triple density measurements in MuSiQue. Results are reported using R@15. Triple densities ($\rho_t$) are calculated as the average number of triples associated with the gold documents for the questions within the MuSiQue's test set.}
\label{tab:robustness_table}
\end{table*}
\section{Qualitative Analysis}
\label{appendix_sec:qualitative_analysis}

\subsection{Positive Instances in MuSiQue}

\begin{table*}[t]
\centering
\small
\begin{tabular}{@{}L{3.5cm}L{6.6cm}L{5cm}@{}}
\toprule
\rowcolor{white} 
\textbf{Query} & \textbf{\gear} & \textbf{HippoRAG w$/$ IRCoT} \\
\midrule

\cellcolor{white}\begin{tabular}[l]{@{}l@{}}What team is the player who\\ led the league in sacks in\\ 2017 a member of?\end{tabular} & 
\cellcolor{customGreen}\begin{tabular}[c]{@{}l@{}}$\langle$Chandler Jones, team, Arizona Cardinals$\rangle$\\ $\langle$Chandler Jones, led the league in sacks in, 2017$\rangle$\end{tabular} & 
\cellcolor{customRed}To answer the question about the player who led the league in sacks in 2017, I need to identify the player and their team from external sources, as the provided documents do not contain that specific information \\
\midrule

\cellcolor{white}\begin{tabular}[l]{@{}l@{}}Who is the CEO of\\ FleetBoston Financial's\\ purchaser?\end{tabular} & 
\cellcolor{customGreen}\begin{tabular}[c]{@{}l@{}}$\langle$FleetBoston Financial, purchaser, Bank of America$\rangle$\\ $\langle$Bank of America, CEO, Brian Moynihan$\rangle$\end{tabular} & 
\cellcolor{customYellow}The purchaser of FleetBoston Financial is Bank of America. To find the CEO of Bank of America, I need to look for information about the current leadership of the bank. Next, I will check for the current CEO of Bank of America to answer the question accurately.\\
\midrule

\cellcolor{white}\begin{tabular}[l]{@{}l@{}}What UK label was \\purchased by the company \\that,  along with the \\distributor of The Cube, \\and ABC, is the other\\ major New York-based\\ broadcaster?\end{tabular} & 
\cellcolor{customGreen}\begin{tabular}[c]{@{}l@{}}$\langle$CBS, purchased, Oriole Records$\rangle$\\ $\langle$CBS, is a major broadcaster based in, New York$\rangle$\\ $\langle$NBC, is a major broadcaster based in, New York$\rangle$\\ $\langle$The Cube, was distributed by, NBC$\rangle$\end{tabular} & 
\cellcolor{customYellow}The original network of Undercovers is NBC, which is one of the major broadcasters based in New York. Now, I need to find out which UK label was bought by NBC ... \\
\midrule

\cellcolor{white}\begin{tabular}[l]{@{}l@{}}What was the 2018\\ population of the Italian\\ city that's underwater?\end{tabular} & 
\cellcolor{customGreen}\begin{tabular}[c]{@{}l@{}}$\langle$Venice, population in 2018, 260\,897$\rangle$\end{tabular} & 
\cellcolor{customRed}The Italian city that is underwater is Krag, British Columbia, which is a ghost town... \\
\bottomrule
\end{tabular}
\caption{Comparison of MuSiQue queries where \gear achieves 100\% recall at R@15 in a single iteration, while HippoRAG w$/$ IRCoT shows lower performance despite using all four available iterations. Cell colors indicate recall performance: \colorbox{customGreen}{green} for 100\% recall, \colorbox{customRed}{red} for 0\% recall, and \colorbox{customYellow}{yellow} for any intermediate value. Cell values in \gear represent the proximal triples stored in the Gist Triple Memory. Cell values in HippoRAG w$/$ IRCoT represent IRCoT's thought process.}
\label{tab:triple_extraction_comparison}
\end{table*}

Table \ref{tab:triple_extraction_comparison} showcases some query instances where \gear achieves perfect recall in a single iteration, while HippoRAG w$/$ IRCoT achieves lower recall and consumes all available iterations. The presented examples illustrate how \gear's Gist Memory $\mathcal{G}^{(n)}$ precisely captures the essential information needed to answer MuSiQue's queries, maintaining the appropriate level of granularity without including superfluous details. In contrast, HippoRAG w/ IRCoT struggles to retrieve crucial information—whether due to limitations in its triple extraction step or retriever functionality—such as the exact population of Venice, which is necessary for accurate responses. Furthermore, the verbose nature of IRCoT's thought process component contrasts with \gear's streamlined approach. The lack of such verbose component in our approach contributes to the fact that \gear requires fewer LLM tokens than the competition, as explained in subsection \ref{subsec:gear_efficient}.

\subsection{Negative Instances in MuSiQue}
We manually assess 20 problematic cases in MuSiQue where \gear did not achieve full recall performance, and we identify the specific error types responsible for each. The findings, presented in Table \ref{tab:error_category_analysis} indicate that the majority of the errors are due to hallucinations of the LLM read steps, and only a very limited number of cases can be attributed to triple extraction.

\begin{table*}[p] 
\centering

\resizebox{\textwidth}{!}{
\begin{tabular}{@{} L{4cm} C{1.9cm} L{13.5cm} @{}} 

\toprule 

\textbf{Error Category} & \textbf{Count} & \textbf{Example} \\

\midrule 

Base Retriever Limitations & 
$2/20$ $ (10\%)$ & 
\textbf{Question:} Who owns the record label where the singer of All Right records? \newline\vspace{0.1cm} 
\textbf{Failure Explanation:} Instead of retrieving passages about the song `All Right,' the base retriever returns passages about the songs `All Right Now' and `Who Owns My Heart,' which are unrelated to the gold passage.\\

\midrule 

Reasoner Hallucinations & $2/20$ $ (10\%)$  &
\textbf{Question:} What is the name of the castle in the city where the headquarters of the production company of A Cosmic Christmas is located? \newline\vspace{0.1cm}
\textbf{Failure Explanation:} The LLM Reasoner concluded the question was answerable, even though key information related to `A Cosmic Christmas' was still missing. \\

\midrule 

Reader Hallucinations & $7/20$ $ (35\%)$ &
\textbf{Question:} Who played the character in Willy Wonka and the Chocolate Factory that the performer of Victrola was named after? \newline\vspace{0.1cm}
\textbf{Failure Explanation:} The triple $\langle$Victrola, named after, Violet Beauregardel$\rangle$ is hallucinated to complete the hop needed to reach the answer. \\

\midrule 

Dataset Issues & $5/20$ $ (25\%)$ &
\textbf{Question:} What other recognition did the Oscar winner for Best Actor in 2006 receive? \newline\vspace{0.1cm}
\textbf{Failure Explanation:} The corpus has no mention that the `Oscars in 2006' were also called the `78th academy awards', preventing the system from finding the relevant information. \\

\midrule 

Missing Details during Triple Extraction & $1/20$ $ (5\%)$ &
\textbf{Question:} What was the form of the language that the last name Sylvester comes from, used in the era of the man crowned Roman Emperor in AD 800, later known as? \newline\vspace{0.1cm}
\textbf{Failure Explanation:} No triple extracted about the last sentence in the gold paragraph (titled `Middle Ages'): `By the reign of Charlemagne, the language had so diverged from the classical that it was later called Medieval Latin.' \\

\midrule 

Quality of the Constructed Graph (Triple Index) & $3/20$ $ (15\%)$ &
\textbf{Question:} What is the largest medical school in the nation where, along with the country of citizenship of the mother of Marie Antoinette, many expelled French Jews relocated? \newline\vspace{0.1cm}
\textbf{Failure Explanation:} The disambiguation between Marie Antoinette and Duchess Marie Antoinette is not correctly handled.\\ 

\bottomrule 

\end{tabular}%

} 
\caption{Analysis of 20 problematic cases in the MuSiQue dataset, divided into various error categories.} 
\label{tab:error_category_analysis} 
\end{table*}

\subsection{Beyond MuSiQuE, 2Wiki and HotpotQA}
\label{appendix_sec:hipporag_data}

We evaluate our framework on three open-domain multi-hop QA datasets: \textbf{MuSiQue} \cite{Trivedi2022}, \textbf{HotpotQA} \cite{Yang2018}, and \textbf{2WikiMultiHopQA} (2Wiki) \cite{Ho2020}. Our dataset choices closely align with the multi-hop QA tasks, and are consistent with related studies in this space \cite{Li2024,Fang2024,Gutierrez2024,Park2024}.

In order to explore the generalisability of \gear in additional scenarios, we use the \textit{hand-picked} case study data\footnote{\url{https://github.com/OSU-NLP-Group/HippoRAG/tree/main/data}} provided by \citeauthor{Gutierrez2024}. These include four path-finding questions across four different domains: books, movies, universities and biomedicine. We test \gear against HippoRAG w$/$ IRCoT on these cases. Table~\ref{tab:case_study_comparison} displays the results. In three out of the four cases, \gear outperforms the competition in recall, successfully identifying more relevant passages, and misses the relevant passages in only one case.

\begin{table*}[thbp]
\centering
\footnotesize
\begin{tabular}{@{}L{2.cm} m{2.9cm} C{0.7cm} C{0.82cm} C{0.9cm} m{2.9cm} C{0.7cm} C{0.82cm} C{0.9cm}@{}}
\toprule
\multirow{2.5}{*}{\textbf{Query}} & 
\multicolumn{4}{c}{\textbf{\gear}} & 
\multicolumn{4}{c}{\textbf{HippoRAG w$/$ IRCoT}} \\
\cmidrule{2-5} \cmidrule{6-9}
& \textbf{Passage Titles} & {R@5} & {R@10} & {R@15} & \textbf{Passage Titles} & {R@5} & {R@10} & {R@15} \\
\midrule
Which book was published in 2012 by an English author who is a Whitbread Award winner? & \parbox{2.88cm}{\(\mathcal{P}_{1}\) A Stitch in Time \\\(\mathcal{P}_{2}\) Stevie Parle\\\(\mathcal{P}_{3}\) \textbf{The Red House}\\\(\mathcal{P}_{4}\) Whitbread Awards\\$\ldots$\\\(\mathcal{P}_{10}\) \textbf{Mark Haddon}}\vfill & 
\cellcolor{customYellow}$50$ & \cellcolor{customGreen}$100$ & \cellcolor{customGreen}$100$ & \parbox{2.88cm}{\(\mathcal{P}_{1}\) Oranges Are Not $\ldots$\\\(\mathcal{P}_{2}\) \textbf{Mark Haddon}\\\(\mathcal{P}_{3}\) William Trevor $\ldots$\\\(\mathcal{P}_{4}\) The Curious $\ldots$\\$\ldots$\\\(\mathcal{P}_{11}\) \textbf{The Red House}}\vfill & 
\cellcolor{customYellow}$50$ & \cellcolor{customYellow}$50$ & \cellcolor{customGreen}$100$ \\
\midrule
Which war film based on a non fiction book was directed by someone famous in the science fiction and crime genres? & 
\parbox{2.88cm}{\(\mathcal{P}_{1}\) And the Band $\ldots$\\\(\mathcal{P}_{2}\) Band of Brothers\\\(\mathcal{P}_{3}\) Aircraft in Fiction\\\(\mathcal{P}_{4}\) Unchained\\$\ldots$}\vfill & 
\cellcolor{customRed}$0$ & \cellcolor{customRed}$0$ & \cellcolor{customRed}$0$ & \parbox{2.88cm}{\(\mathcal{P}_{1}\) Shangai Patrol\\\(\mathcal{P}_{2}\) \textbf{Black Hawk Down}\\\(\mathcal{P}_{3}\) \textbf{Ridley Scott}\\\(\mathcal{P}_{4}\) Outline of science \\$\ldots$}\vfill & 
\cellcolor{customGreen}$100$ & \cellcolor{customGreen}100 & \cellcolor{customGreen}$100$ \\
\midrule
Which Stanford professor works on the neuroscience of Alzheimer's? & 
\parbox{2.88cm}{\(\mathcal{P}_{1}\) Thomas C. Sudhof\\\(\mathcal{P}_{2}\) \textbf{Thomas C. Sudhof}\\\(\mathcal{P}_{3}\) Judes Poirier\\\(\mathcal{P}_{4}\) \textbf{Thomas C. Sudhof}\\$\ldots$ \\\(\mathcal{P}_{10}\) \textbf{Robert Malenka} \\\(\mathcal{P}_{13}\) \textbf{Robert Malenka}}\vfill & 
\cellcolor{customYellow}$50$ & \cellcolor{customYellow}$75$ & \cellcolor{customGreen}$100$ & 
\parbox{2.88cm}{\(\mathcal{P}_{1}\) \textbf{Thomas C. Sudhof}\\\(\mathcal{P}_{2}\) \textbf{Thomas C. Sudhof}\\\(\mathcal{P}_{3}\) Manolis Kellis\\\(\mathcal{P}_{4}\) Giovanna Malluci\\\(\mathcal{P}_{5}\) Dena Dubal}\vfill & 
\cellcolor{customYellow}$50$ & \cellcolor{customYellow}$50$ & \cellcolor{customYellow}$50$ \\
\midrule
What drug is used to treat chronic lymphocytic leukemia by interacting with cytosolic p53? & 
\parbox{2.88cm}{\(\mathcal{P}_{1}\) P53 Regulation\\\(\mathcal{P}_{2}\) Venetoclax\\\(\mathcal{P}_{3}\) \textbf{Chlorambucil}\\\(\mathcal{P}_{4}\) Chronic Lymphocytic Leukemia}\vfill & 
\cellcolor{customYellow}$50$ & \cellcolor{customYellow}$50$ & \cellcolor{customYellow}$50$ & 
\parbox{2.88cm}{\(\mathcal{P}_{1}\) P53\\\(\mathcal{P}_{2}\) Cirmtuzumab\\\(\mathcal{P}_{3}\) MDC1 Function\\\(\mathcal{P}_{4}\) Chronic Lymphocytic Leukemia}\vfill & 
\cellcolor{customRed}$0$ & \cellcolor{customRed}$0$ & \cellcolor{customRed}$0$ \\
\bottomrule
\end{tabular}
\caption{Retrieval performance comparison between \gear and HippoRAG w$/$ IRCoT. Both models are configured with a maximum number of 4 iterations. The example questions are taken from \citeauthor{Gutierrez2024} and showcase multi-hop path-finding queries across different domains: books, movies, universities and biomedicine. Cell colours indicate recall performance: \colorbox{customGreen}{green} for 100\% recall, \colorbox{customRed}{red} for 0\% recall, and \colorbox{customYellow}{yellow} for any intermediate value. Retrieved passage titles are listed in the 'Passage Titles' columns, with \textbf{bold} text indicating gold passages and $\mathcal{P}_{n}$ indicating their position in the retrieved list.}
\label{tab:case_study_comparison}
\end{table*}

\section{Increasing the Number of Agent Iterations}
\label{appendix_sec:increasing_n_iterations}
\begin{figure*}[p]
\centering
\includegraphics[width=\textwidth]{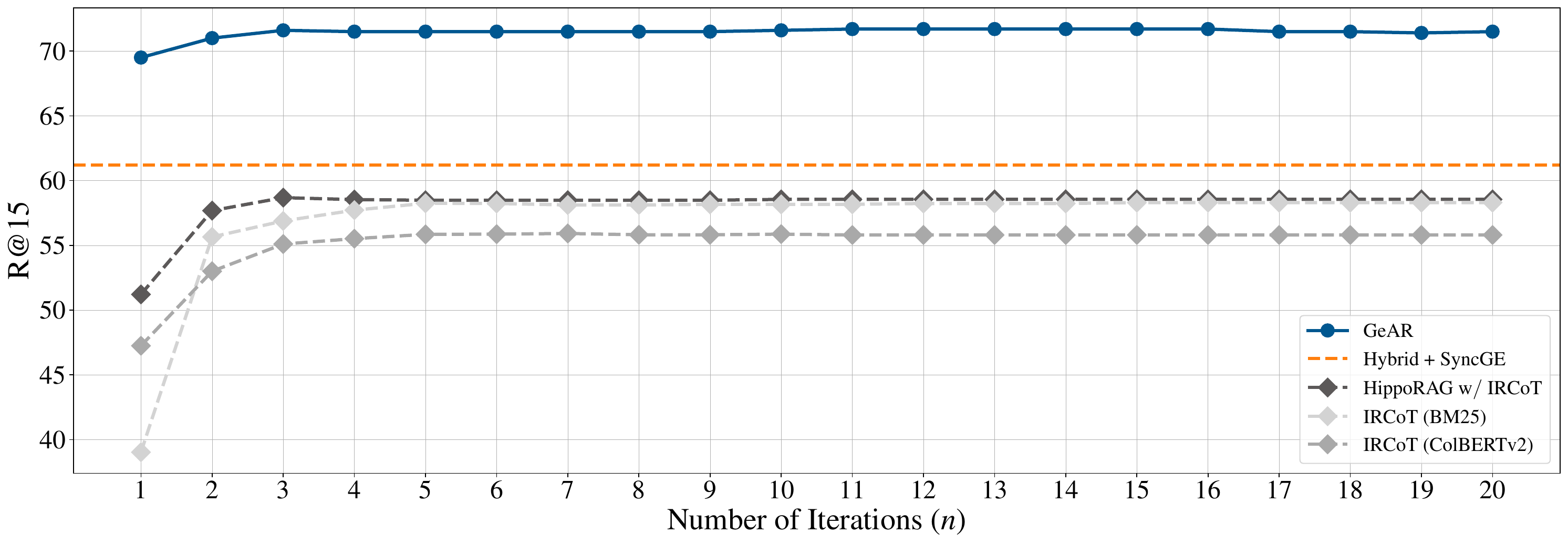}
\caption{Evolution of R@15 over 20 iterations on MuSiQue. Recall is computed at each iteration using the cumulative set of retrieved documents, with prior recall values carried forward for questions that terminated in earlier iterations. The horizontal line indicates the single-step performance of Hybrid + SyncGE.}
\label{fig:recall_across_iterations_20_iters}
\end{figure*}

Figure \ref{fig:recall_across_iterations_20_iters} expands upon the analysis shown in Figure \ref{fig:recall_across_iterations} by evaluating retrieval performance over 20 iterations, rather than the initial 4 iterations. The results demonstrate a consistent pattern across all methods: retrieval performance stabilises after approximately 4 iterations, with no substantial improvements or degradation in subsequent iterations. While some minor fluctuations occur beyond this point, they are negligible.

This performance plateau can be attributed to two key factors. First, the query re-writing mechanisms in all investigated approaches struggle to generate effective subsequent queries. Second, our analysis has identified several cases of unanswerable queries within MuSiQue's answerable subset. A representative example is provided in Table~\ref{tab:musique_problematic_example}.

\begin{table*}
\centering
\footnotesize
\begin{tabular}{L{3cm}L{12cm}}

\toprule
\textbf{Question} & Who did the \textcolor{red}{producer} of \textcolor{purple}{Big Jim McLain} play in \textcolor{purple}{True Grit}? \\
\midrule
\multirow{3.5}{*}{\textbf{Gold Passages}} & {1. \textcolor{purple}{Big Jim McLain}: \textcolor{purple}{Big Jim McLain} is a 1952 political thriller film starring John Wayne and James Arness as HUAC investigators.}\\\cmidrule{2-2}
& {2. \textcolor{purple}{True Grit} is a 1969 American western film. It is the first film adaptation of Charles Portis' 1968 novel of the same name. The screenplay was written by Marguerite Roberts. The film was directed by Henry Hathaway and starred Kim Darby as Mattie Ross and John Wayne as U.S. Marshal Rooster Cogburn. Wayne won his only Academy Award for his performance in this film and reprised his role for the 1975 sequel Rooster Cogburn.}\\
\midrule
\textbf{Comment} & \textcolor{red}{No information about who was the producer of} \textcolor{purple}{Big Jim McLain} \textcolor{red}{is provided in the gold passages}\\
\bottomrule
\end{tabular}
\caption{\label{tab:musique_problematic_example}Example of a query from MuSiQue that is not answerable solely based on the provided gold passages.}
\end{table*}

\clearpage
\onecolumn  
\section{Prompts}
\label{appendix_sec:agent_prompts}

\subsection{Offline Prompts}
\label{sec:offline_prompts}

\begingroup\hypersetup{linkcolor=white}
\begin{prompt}[title={Triple Extraction\hfill{}(\S\ref{subsec:implementation_details}})]
 \textbf{\# Instruction} \\
\\
Your task is to construct an RDF (Resource Description Framework) graph from the given passages and named entity lists. \\
Respond with a JSON list of triples, with each triple representing a relationship in the RDF graph. \\
Pay attention to the following requirements: \\
- Each triple should contain at least one, but preferably two, of the named entities in the list for each passage. \\
- Clearly resolve pronouns to their specific names to maintain clarity. \\
\\
Convert the paragraph into a JSON dict containing a named entity list and a triple list. \\
\\
\ \textbf{\# Demonstration \#1} \\
\\
Paragraph: \\
``` \\
Magic Johnson \\
\\
After winning a national championship with Michigan State in 1979, Johnson was selected first overall in the 1979 NBA draft by the Lakers, leading the team to five NBA championships during their "Showtime" era. \\
``` \\
\{\{"named\_entities": ["Michigan State", "national championship", "1979", "Magic Johnson", \\ "National Basketball Association", "Los Angeles Lakers", "NBA Championship"]\}\} \\
\{\{ \\
    "triples": [ \\
        ("Magic Johnson", "member of sports team", "Michigan State"), \\
        ("Michigan State", "award", "national championship"), \\
        ("Michigan State", "award date", "1979"), \\
        ("Magic Johnson", "draft pick number", "1"), \\
        ("Magic Johnson", "drafted in", "1979"), \\
        ("Magic Johnson", "drafted by", "Los Angeles Lakers"), \\
        ("Magic Johnson", "member of sports team", "Los Angeles Lakers"), \\
        ("Magic Johnson", "league", "National Basketball Association"), \\
        ("Los Angeles Lakers", "league", "National Basketball Association"), \\
        ("Los Angeles Lakers", "award received", "NBA Championship"), \\
    ] \\
\}\} \\
``` \\
\\
\ \textbf{\# Demonstration \#2} \\
\\
Paragraph: \\
``` \\
Elden Ring \\
\\
Elden Ring is a 2022 action role-playing game developed by FromSoftware. It was directed by Hidetaka Miyazaki with worldbuilding provided by American fantasy writer George R. R. Martin. \\
``` \\
\{\{"named\_entities": ["Elden Ring", "2022", "Role-playing video game", "FromSoftware", "Hidetaka Miyazaki", "United States of America", "fantasy", "George R. R. Martin"]\}\} \\
\{\{ \\
    "triples": [ \\
        ("Elden Ring", "publication", "2022"), \\
        ("Elden Ring", "genre", "action role-playing game"), \\
        ("Elden Ring", "publisher", "FromSoftware"), \\
        ("Elden Ring", "director", "Hidetaka Miyazaki"), \\
        ("Elden Ring", "screenwriter", "George R. R. Martin"), \\
        ("George R. R. Martin", "country of citizenship", "United States of America"), \\
        ("George R. R. Martin", "genre", "fantasy"), \\
    ] \\
\}\} \\
\\
\\
\ \textbf{\# Input} \\
\\
Convert the paragraph into a JSON dict, it has a named entity list and a triple list. \\
\\
Paragraph: \\
``` \\
\textbf{$\{$wiki\_title$\}$} \\
\\
\textbf{$\{$passage$\}$}\\
\end{prompt}
\endgroup

\subsection{Online Retrieval Prompts}
\label{subsec:online_retrieval_prompts}

The \textcolor{blue}{blue}-highlighted portions of the Reader prompt below indicate additional text that is only required when the Gist Memory $\mathcal{G}^{(n)}$ is active. When Gist Memory is inactive, these blue sections should be omitted, and the $\{$triples$\}$ parameter should be left empty.

\begingroup\hypersetup{linkcolor=white}
\begin{prompt}[title={Reader with and without Gist Memory\hfill{}(\S\ref{subsection:knowledge_syncro}} and \S\ref{subsec:gist_memory_constructor})]
Your task is to find facts that help answer an input question. \\
\\
You should present these facts as knowlege triples, which are structured as ("subject", "predicate", "object"). \\
Example: \\
Question: When was Neville A. Stanton's employer founded? \\
Facts: ("Neville A. Stanton", "employer", "University of Southampton"), ("University of Southampton", "founded in", "1862") \\
\\
\\
Now you are given some documents:\\
\textbf{$\{$docs$\}$} \\
\\
\\
Based on these documents \textcolor{blue}{and some preliminary facts provided below}, \\ find additional supporting fact(s) that may help answer the following question. \\
 \\
Note: if the information you are given is insufficient, output only the relevant facts you can find.\\
\\
Question: \textbf{$\{$query$\}$} \\
Facts: \textcolor{blue}{\textbf{$\{$triples$\}$}} \\
\end{prompt}
\endgroup

\begingroup\hypersetup{linkcolor=white}
\begin{prompt}[title={Reasoning for Termination\hfill{}(\S\ref{subsec:reasoning}})]
\ \textbf{\# Task Description:} \\
You are given an input question and a set of known facts:\\
Question: \textbf{$\{$query$\}$} \\
Facts: \textbf{$\{$triples$\}$} \\
\\
Your reply must follow the required format:\\
1. If the provided facts contain the answer to the question, your should reply as follows:\\
Answerable: Yes\\
Answer: ...\\
\\
2. If not, you should explain why and reply as follows:\\
Answerable: No\\
Why: ...\\
\\
\ \textbf{\# Your reply:} \\
\end{prompt}
\endgroup

\begingroup\hypersetup{linkcolor=white}
\begin{prompt}[title={Query Re-writing\hfill{}(\S\ref{subsec:rewriting}})]
\ \textbf{\# Task Description:} \\
You will be presented with an input question and a set of known facts. \\
These facts might be insufficient for answering the question for some reason. \\
Your task is to analyze the question given the provided facts and 
determine what else information is needed for the next step. \\
\\
\ \textbf{\# Example:} \\
Question: What region of the state where Guy Shepherdson was born, contains SMA Negeri 68?\\
Facts: ("Guy Shepherdson", "born in", "Jakarta")\\
Reason: The provided facts only indicate that Guy Shepherdson was born in Jakarta, but they do not provide any information about the region of the state that contains SMA Negeri 68. \\
Next Question: What region of Jakarta contains SMA Negeri 68? \\
\\
\ \textbf{\# Your Task:} \\
Question: \textbf{$\{$query$\}$} \\
Facts: \textbf{$\{$triples$\}$} \\
Reason: \textbf{$\{$reason$\}$} \\
\\
Next Question:
\end{prompt}
\endgroup

\subsection{Online Question Answering Prompts}

The following prompt with retrieved passages combines the QA generation prompts from \citeauthor{Gutierrez2024} and \citeauthor{Wang2024}. For the variation without retrieved passages, we omit the preamble and only include the instruction, highlighted in \textcolor{purple}{purple} .

\begingroup\hypersetup{linkcolor=white}
\begin{prompt}[title={Retrieved Passages with In-context Example\hfill{}(Table \ref{tab:em_and_f1}})]

As an advanced reading comprehension assistant, your task is to analyze text passages and corresponding questions meticulously, with the aim of providing the correct answer. \\
==================\\
For example:\\
==================\\
Wikipedia Title: Edward L. Cahn \\
Edward L. Cahn (February 12, 1899 – August 25, 1963) was an American film director.\\
\\
Wikipedia Title: Laughter in Hell \\
Laughter in Hell is a 1933 American Pre-Code drama film directed by Edward L. Cahn and starring Pat O'Brien. The film's title was typical of the sensationalistic titles of many Pre-Code films. Adapted from the 1932 novel of the same name buy Jim Tully, the film was inspired in part by "I Am a Fugitive from a Chain Gang" and was part of a series of films depicting men in chain gangs following the success of that film. O'Brien plays a railroad engineer who kills his wife and her lover in a jealous rage and is sent to prison. The movie received a mixed review in "The New York Times" upon its release. Although long considered lost, the film was recently preserved and was screened at the American Cinematheque in Hollywood, CA in October 2012. The dead man's brother ends up being the warden of the prison and subjects O'Brien's character to significant abuse. O'Brien and several other characters revolt, killing the warden and escaping from the prison. The film drew controversy for its lynching scene where several black men were hanged. Contrary to reports, only blacks were hung in this scene, though the actual executions occurred off-camera (we see instead reaction shots of the guards and other prisoners). The "New Age" (an African American weekly newspaper) film critic praised the scene for being courageous enough to depict the atrocities that were occurring in some southern states. \\
\\
Wikipedia Title: Theodred II (Bishop of Elmham) \\
Theodred II was a medieval Bishop of Elmham. The date of Theodred's consecration unknown, but the date of his death was sometime between 995 and 997. \\
\\
Wikipedia Title: Etan Boritzer \\
Etan Boritzer( born 1950) is an American writer of children 's literature who is best known for his book" What is God?" first published in 1989. His best selling" What is?" illustrated children's book series on character education and difficult subjects for children is a popular teaching guide for parents, teachers and child- life professionals. Boritzer gained national critical acclaim after" What is God?" was published in 1989 although the book has caused controversy from religious fundamentalists for its universalist views. The other current books in the" What is?" series include What is Love?, What is Death?, What is Beautiful?, What is Funny?, What is Right?, What is Peace?, What is Money?, What is Dreaming?, What is a Friend?, What is True?, What is a Family?, What is a Feeling?" The series is now also translated into 15 languages. Boritzer was first published in 1963 at the age of 13 when he wrote an essay in his English class at Wade Junior High School in the Bronx, New York on the assassination of John F. Kennedy. His essay was included in a special anthology by New York City public school children compiled and published by the New York City Department of Education. \\
\\
Wikipedia Title: Peter Levin \\
Peter Levin is an American director of film, television and theatre. \\
\\
Question: When did the director of film Laughter In Hell die? \\
Answer: August 25, 1963. \\
================== \\
\textcolor{purple}{Given the following text passages and questions, please present a concise, definitive answer, devoid of additional elaborations, and of maximum length of 6 words.} \\
================== \\
\\
Wikipedia Title : \textbf{$\{$title$\}$}
\textbf{$\{$text$\}$} \texttt{for each retrieved passage} ...  \\
Question: \textbf{$\{$question$\}$} \\
\\
Answer:
\end{prompt}
\endgroup

\label{subsec:online_qa_prompts}

\begingroup\hypersetup{linkcolor=white}
\begin{prompt}[title={No Retrieved Passages\hfill{}(Table \ref{tab:em_and_f1}})]
\textcolor{purple}{Given the following question, please present a concise, definitive answer, devoid of additional elaborations, and of maximum length of 6 words.} \\
\\
Question: \textbf{$\{$question$\}$} \\
\\
Answer:
\end{prompt}
\endgroup

\end{document}